\documentclass[10pt,twocolumn,letterpaper]{article}

\usepackage{iccv}
\usepackage{times}
\usepackage{epsfig}
\usepackage{graphicx}
\usepackage{amsmath}
\usepackage{amssymb}
\usepackage{multirow}
\usepackage{tikz}
\usetikzlibrary{positioning}

\usepackage[utf8]{inputenc}
\usepackage{polski}

\usepackage{nicefrac}       
\usepackage{microtype}      
\usepackage{xcolor}         

\usepackage{url}            

\usepackage{wrapfig}

\def\R{\mathbb{R}}

\def\x{ {\bf x} }
\def\p{ {\bf p} }
\def\q{ {\bf q} }

\def\L{\mathcal{L}}
\def\F{\mathcal{F}}
\def\LP{\mathcal{LP}}

\usepackage[breaklinks=true,bookmarks=false]{hyperref}

\iccvfinalcopy 

\ificcvfinal\fi

\def\our{NeRFlame}

\def\L{\mathcal{L}}

\begin{document}

\title{ \our{}: FLAME-based conditioning of NeRF for 3D face rendering}

\author{Wojciech Zaj\k{a}c, Joanna Waczyńska, Piotr Borycki, Jacek Tabor\\
Faculty of Mathematics and Computer
Science, Jagiellonian University,Poland\\
\and
Maciej Zięba\\
Department of Artificial Intelligence, University of Science and Technology 
Wrocław, Poland\\
\and
Przemysław Spurek\\
Faculty of Mathematics and Computer
Science, Jagiellonian University, Poland\\
{\tt\small  przemylaw.spurek@uj.edu.pl}
}

\maketitle
\ificcvfinal\thispagestyle{empty}\fi


\begin{abstract}
{\small
Traditional 3D face models are based on mesh representations with texture. One of the most important models is FLAME (Faces Learned with an Articulated Model and Expressions), which produces meshes of human faces that are fully controllable. 
Unfortunately, such models have problems with capturing geometric and appearance details. 
In contrast to mesh representation, the neural radiance field  (NeRF) produces extremely sharp renders. However, implicit methods are hard to animate and do not generalize well to unseen expressions. It is not trivial to effectively control NeRF models to obtain face manipulation. 

The present paper proposes a novel approach, named \our{}, which combines the strengths of both NeRF and FLAME methods. Our method enables high-quality rendering capabilities of NeRF while also offering complete control over the visual appearance, similar to FLAME.

In contrast to traditional NeRF-based structures that use neural networks for RGB color and volume density modeling, our approach utilizes the FLAME mesh as a distinct density volume. Consequently, color values exist only in the vicinity of the FLAME mesh. This FLAME framework is seamlessly incorporated into the NeRF architecture for predicting RGB colors, enabling our model to explicitly represent volume density and implicitly capture RGB colors.          
}
\end{abstract}

\begin{figure}[h]
    \begin{tabular}{@{}c@{}c@{}c@{}}
	Original position & \multicolumn{2}{c}{ Modified positions } \\
    \includegraphics[width=0.16\textwidth, trim=50 50 50 50, clip]{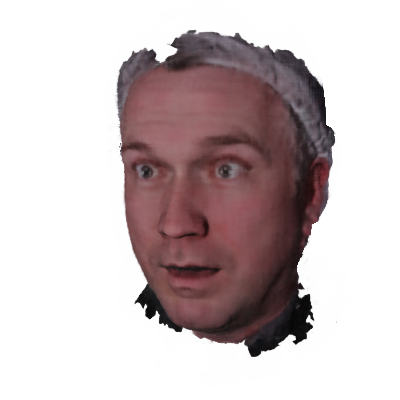}
    &
    \includegraphics[width=0.16\textwidth, trim=50 50 50 50, clip]{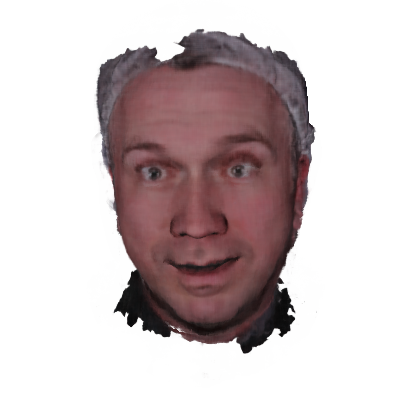} & 
    \includegraphics[width=0.16\textwidth, trim=50 50 50 50, clip]{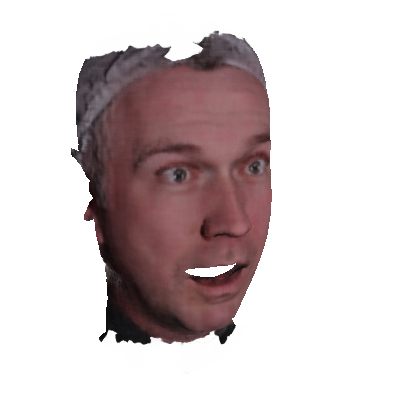} 
    \\ 

    \includegraphics[width=0.16\textwidth, trim=50 50 50 50, clip]{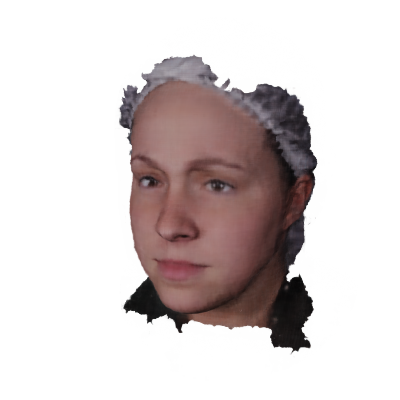}
    &
    \includegraphics[width=0.16\textwidth, trim=50 50 50 50, clip]{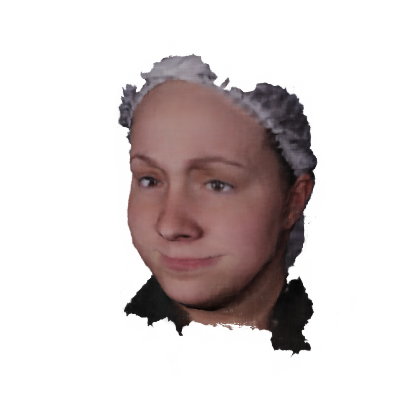} &
    \includegraphics[width=0.16\textwidth, trim=50 50 50 50, clip]{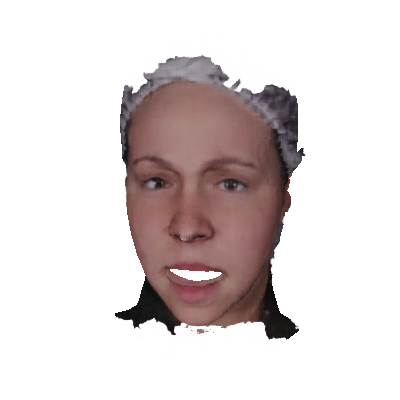} \\
	\end{tabular}
\caption{ Our model facilitates the manipulation of facial attributes in the context of human visage. \our{} use FLAME as a conditioning factor in the NeRF-based model. After training in one fixed position with many views, we can use Flame conditioning to manipulate the face to obtain new facial expressions. We can produce novel views in training positions as well as in modified facial expressions.  }
\label{fig:FLAME_compare_tezer} 
\end{figure}

\section{Introduction}

Methods to automatically create fully controllable human face avatars have many applications in VR/AR and games. Traditional 3D face models are based on fully controllable mesh representations. FLAME~\cite{li2017learning} is one of the most important methods for mesh-based avatars. FLAME integrates a linear shape space trained using 3800 human head scans with articulated jaw, neck, eyeballs, pose-dependent corrective blend shapes, and extra global expression blend shapes. In practice, we can easily train FLAME on the 3D scan (or 2D image) of human faces and then manipulate basic behaviors like jaw, neck, and eyeballs. We can also produce colors for mesh by using textures. Unfortunately, such models have problems with capturing geometric and appearance details.

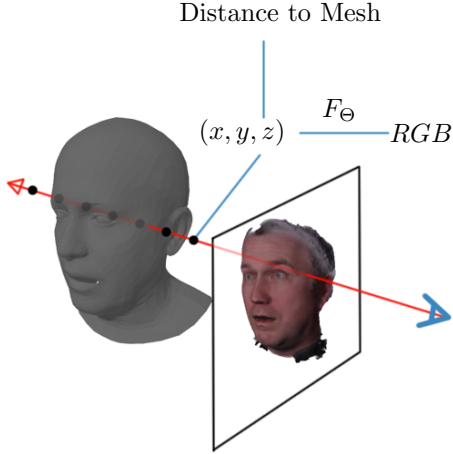
\begin{figure}[!h]
    \begin{center}
    \begin{tikzpicture}[scale=1.0]
    \node[inner sep=0pt] (russell) at (0,0)
    {\includegraphics[width=0.49\textwidth]{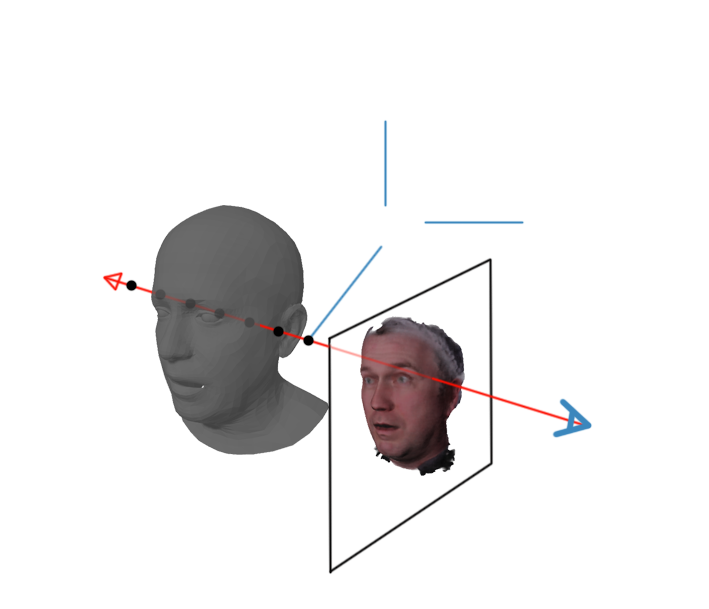} };
    \node[text width=3.5cm] at (1.3,1.0) {$ (x,y,z) $};
    \node[text width=3.5cm] at (3.8,1.0) {$ RGB $};
    \node[text width=3.5cm] at (2.9,1.3) {$ 
    F_{\Theta}$};
    \node[text width=3.5cm] at (1.0,2.6) {Distance to Mesh};
    \end{tikzpicture}
    \caption{\our{} architecture uses two main components: FLAME model and NeRF-based implicit representation. For position $(x,y,z)$, we use distance to the mesh for conditioning volume density $\sigma$ and NeRF architecture to obtain color. In practice \our{} takes Mesh produced by Flame and uses distance to triangle mesh to produce density and color value.
    } 
    \label{fig:architecture}
    \end{center}
\end{figure}

In contrast to the classical approaches, we can use implicit methods that represent avatars using neural networks. NeRFs~\cite{mildenhall2020nerf} represent a scene using a fully connected architecture. As input, they take a 5D coordinate (spatial location $ \x = (x, y, z)$ and viewing direction ${\bf d} =  (\theta, \Psi)$) and return an emitted color ${\bf c} = (r, g, b)$ and volume density $\sigma$.
NeRF extracts information from unlabelled 2D views to obtain 3D shapes. NeRF allows for the synthesizing of novel views of complex 3D scenes from a small subset of 2D images. Based on the relations between those base images and computer graphics principles, such as ray tracing, this neural network model can render high-quality images of 3D objects from previously unseen viewpoints. 
In contrast to the mesh representation, NeRF captures geometric and appearance details. However, it is not trivial to effectively control NeRF to obtain face manipulation. 

There are many approaches to controlling NeRF, which use generative models~\cite{kania2023hypernerfgan,chan2022efficient,zimny2022points2nerf}, dynamic scene encoding~\cite{kania2022conerf}, or conditioning mechanisms~\cite{athar2021flame}. However, our ability to manipulate NeRF falls short compared to our proficiency in controlling mesh representations.

This paper proposes \our{}\footnote{ The source code is available at: \url{https://github.com/WojtekZ4/NeRFlame}.}, a hybrid approach for 3D face rendering that uses implicit and explicit representations; see Fig.~\ref{fig:FLAME_compare_tezer},
Fig.~\ref{fig:architecture} shows the main concept of the model, which is based on two components: NERF and mesh, with only points in the mesh surroundings treated as nerf inputs. Our method inherits the best features from the above approaches by modeling the quality of NeRF rendering and controlling the appearance as in FLAME. We combine those two techniques by showing how to condition the NeRF model by mesh effectively. Our model's fundamental idea is conditioning volume density by distance to mesh. The volume density is non-zero only in the $\varepsilon$ neighborhood of FLAME mesh. 
Therefore, we use the NeRF-based architecture to model volume density and RGB colors only in the $\varepsilon$ neighborhood of mesh. Such a solution allows for obtaining similar quality renders to NeRF and a level of controlling mesh similar to FLAME (table \ref{tab:nerf}).
In contrast to Dynamic Neural Radiance Fields, \our{}  undergoes training using a single position of the human face rather than relying on sequences from various positions in movies. Despite this distinction, our model exhibits comparable functionality. We have the capability to generate a range of facial expressions and novel perspectives for newly encountered face positions facilitated by the FLAME backbone. This enables us to model previously unseen facial expressions. Consequently, we conduct a comparison between our model and the traditional static Neural Radiance Fields.


The contributions of this paper are significant and are outlined as follows:
\begin{itemize}
    \item We introduce \our{} -- an innovative  NeRF model conditioning be  FLAME that combines the best features of both methods, namely the exceptional rendering quality of NeRF and the precise control over appearance as in FLAME.
    \item We demonstrate the ability to condition model volume density in NeRF by employing mesh representation, which represents a significant advancement over traditional NeRF-based approaches that rely on neural networks. 
    \item We train our model on a single position of the human face rather than using entire movies, thereby highlighting the versatility and practicality of our approach.
\end{itemize}
Overall, our contributions offer a substantial advancement in the realms of 3D facial modeling and rendering, providing a foundation for future exploration and research in this domain.

\begin{figure*}[h]
	\centering
    \begin{tabular}{@{}c@{}c@{}c@{}c@{}c@{}c@{}c@{}}
	Original views & \multicolumn{2}{c}{  \our{} } & \multicolumn{2}{c}{ Classical FLAME }\\
\includegraphics[width=0.19\textwidth, trim=100 100 100 100, clip]{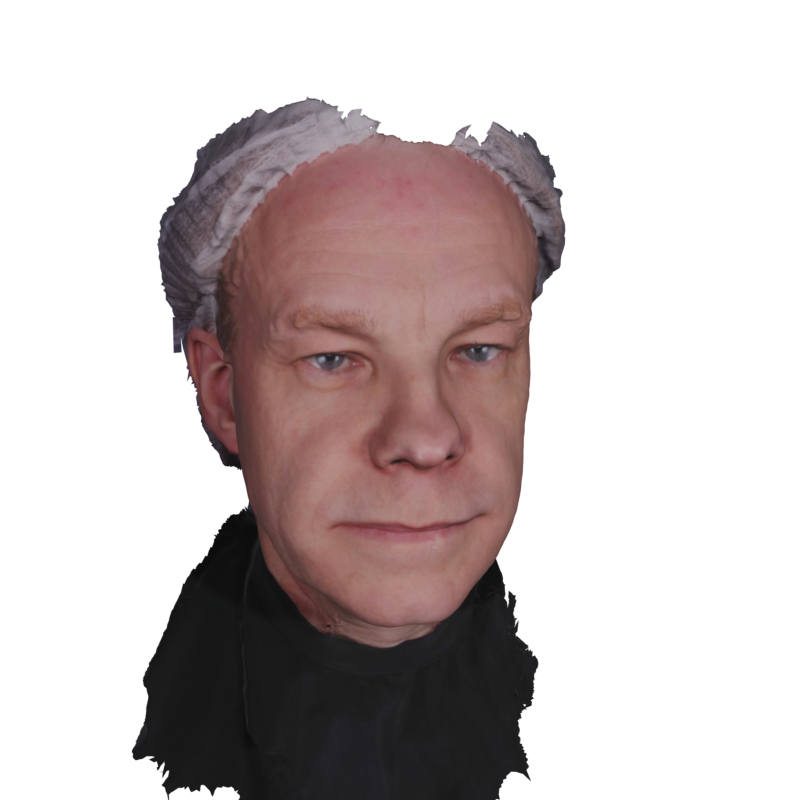}
    &
    \includegraphics[width=0.19\textwidth, trim=50 50 50 50, clip]{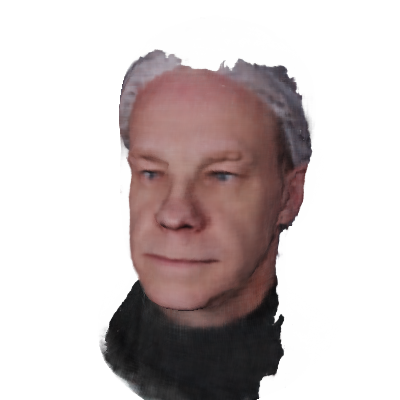} &
    \includegraphics[width=0.19\textwidth, trim=50 50 50 50, clip]{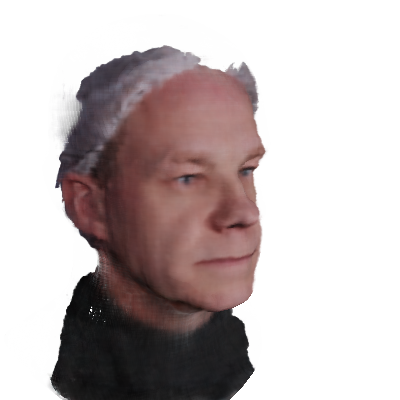}
    &
    \includegraphics[width=0.19\textwidth, trim=100 100 100 100, clip]{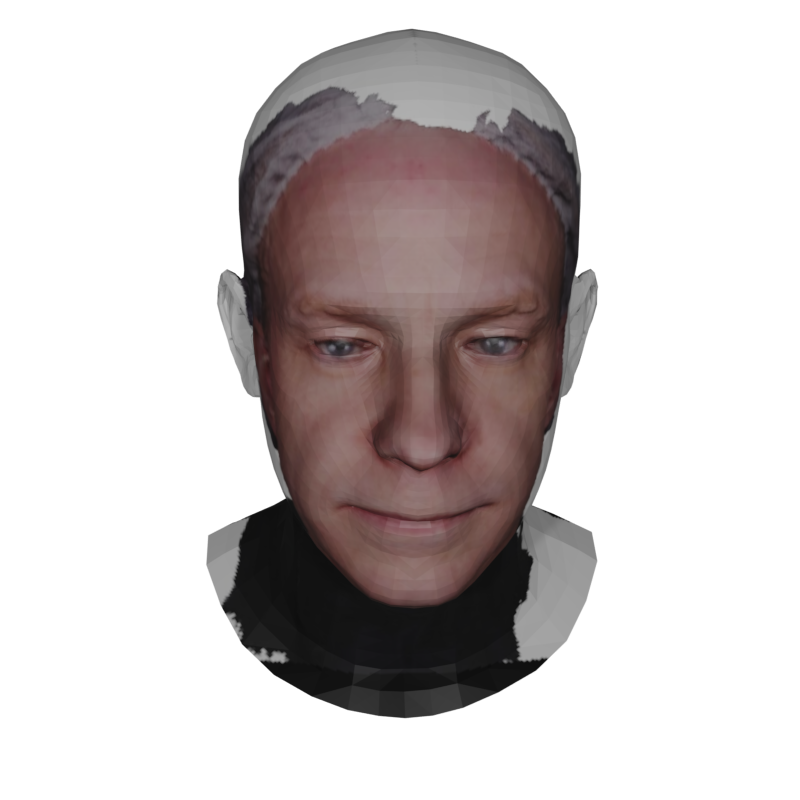} &
    \includegraphics[width=0.19\textwidth, trim=100 100 100 100, clip]{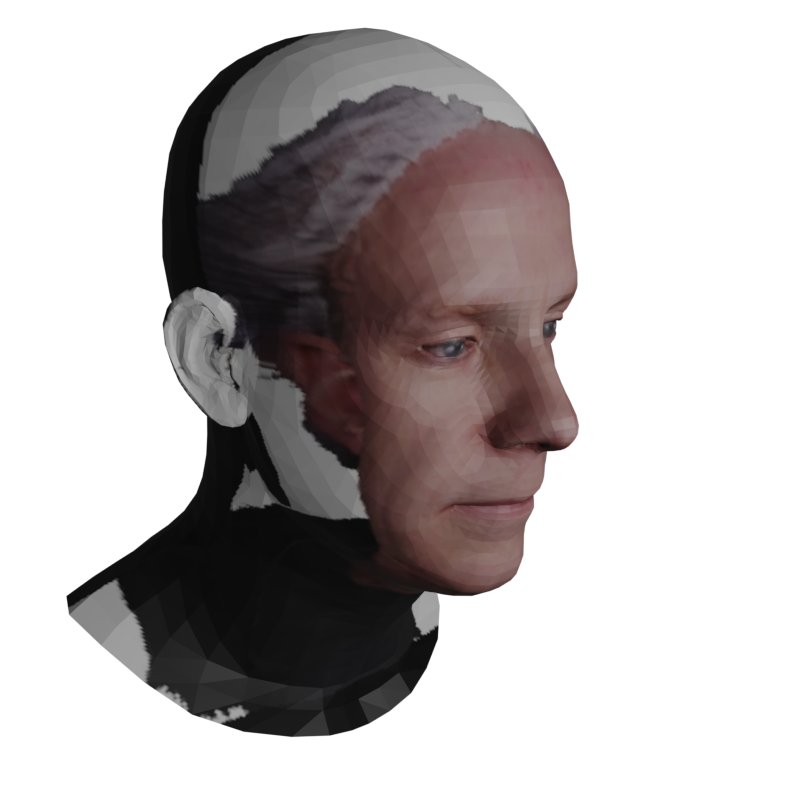}\\
\includegraphics[width=0.19\textwidth, trim=100 100 100 100, clip]{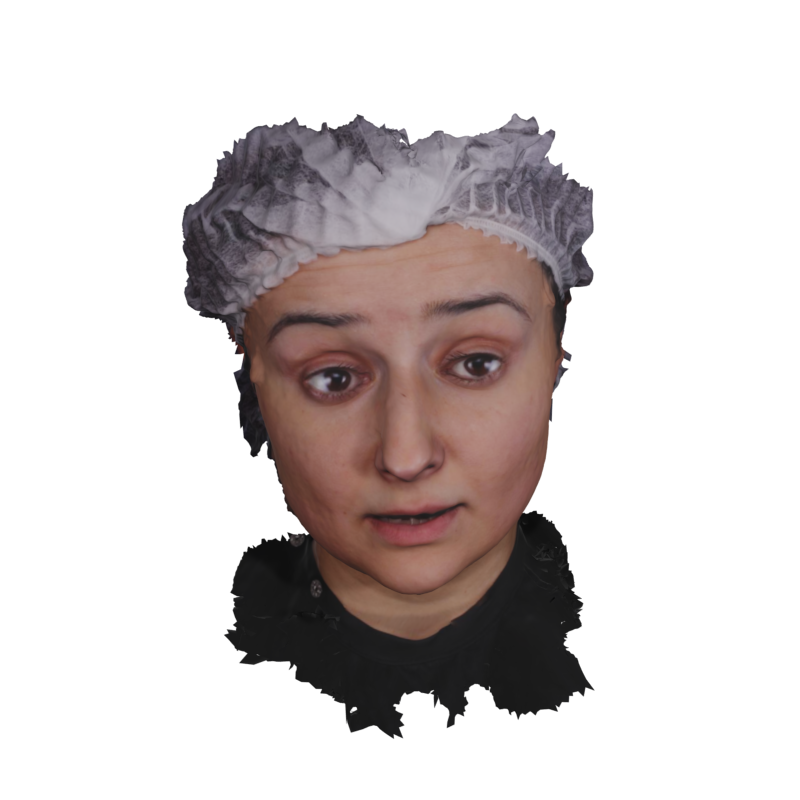}
    & 
    \includegraphics[width=0.19\textwidth, trim=50 50 50 50, clip]{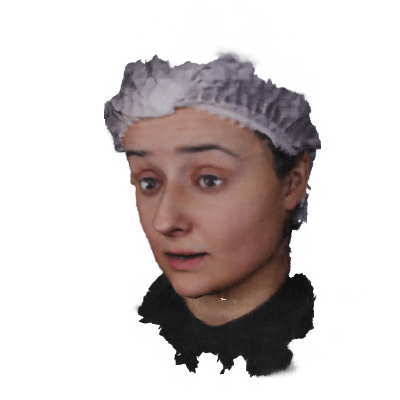} &
    \includegraphics[width=0.19\textwidth, trim=50 50 50 50, clip]{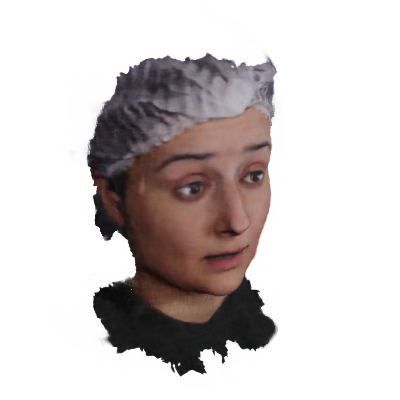} &

    \includegraphics[width=0.19\textwidth, trim=100 150 100 50, clip]{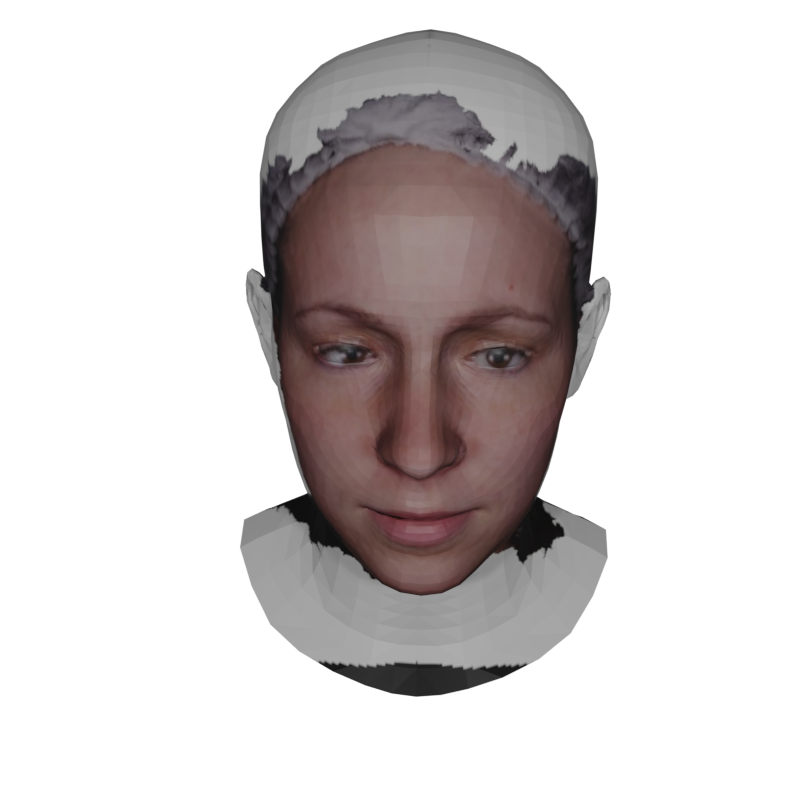} &
    \includegraphics[width=0.19\textwidth, trim=100 150 100 50, clip]{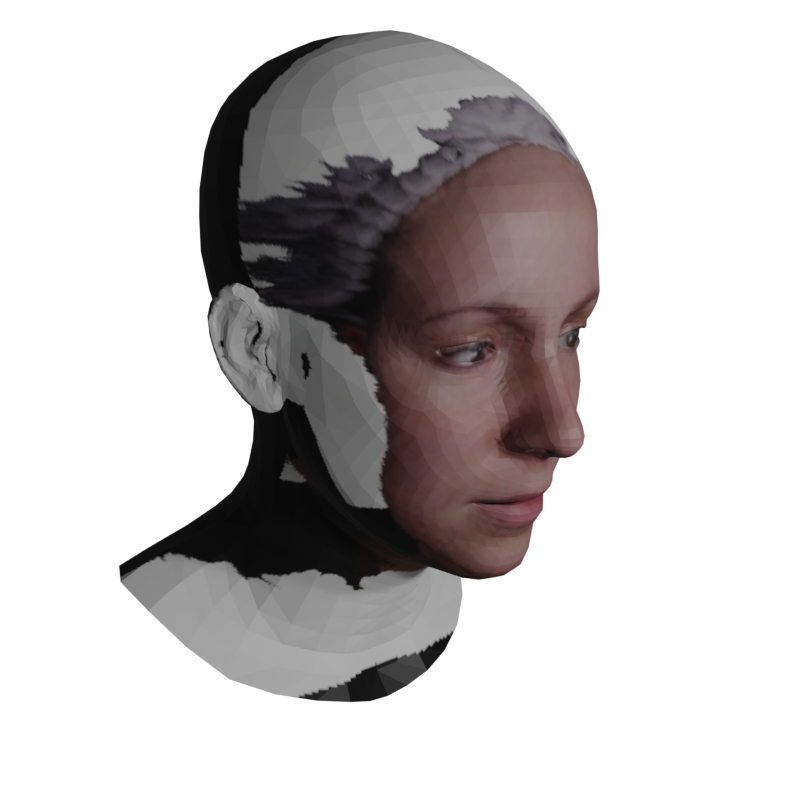}\\
  
	\end{tabular}
\caption{Competition between \our{} and classical FLAME fitting. NeRF-based model better fits human expression of the face. }
\label{fig:FLAME_compare} 
\end{figure*}

\section{Related Works}

\our{} is a model of controllable human face avatars trained on a single 3D face represented by a few 2D images. Given our model's training on 2D facial images, we naturally refer to Static Neural Radiance Fields. However, our ability to adapt NeRF is noteworthy due to the utilization of FLAME as a backbone. Consequently, in our exploration of related works, we incorporate considerations for Dynamic Neural Radiance Fields.

\paragraph{Static Neural Radiance Fields} 
3D objects can be represented by using many different approaches, including voxel grids~\cite{choy20163d}, octrees \cite{hane2017hierarchical}, multi-view images \cite{arsalan2017synthesizing,LIU2022108774}, point clouds \cite{achlioptas2018learning,shu2022wasserstein,yang2022continuous}, geometry
images \cite{sinha2016deep}, deformable meshes \cite{girdhar2016learning,li2017learning},
and part-based structural graphs \cite{li2017grass}.

The above representations are discreet, which causes some problems in real-life applications. In contrast to such apprehension, NeRF~\cite{mildenhall2020nerf} represents a scene using a fully-connected architecture. NeRF and many generalizations \cite{barron2021mip,barron2022mip,liu2020neural,niemeyer2022regnerf,roessle2022dense,tancik2022block,verbin2022ref}  synthesize novel views of a static scene using differentiable volumetric rendering.

One of the largest limitations is training time. To solve such problems in \cite{fridovich2022plenoxels}, authors propose Plenoxels, a method that uses
a sparse voxel grid storing density and spherical harmonics coefficients at each node. The final color is the composition of tri-linearly interpolated values of each voxel. In \cite{muller2022instant}, authors use a similar approach, but the space is divided into an independent multilevel grid.
In \cite{chen2022tensorf}, authors represent a 3D object as an orthogonal tensor component. A small MLP network, which uses orthogonal projection on tensors, obtains the final color and density. 
Some methods use additional information to Nerf, like depth maps or point clouds
\cite{azinovic2022neural,deng2022depth,roessle2022dense,wei2021nerfingmvs}.

In our paper, we produce a new NeRF-based representation of 3D objects. As input, we use classical 2D images. However, RGB colors and volume density are conditioned by distance to FLAME mesh.

\paragraph{Dynamic Neural Radiance Fields}

Current solutions for implicit reprehension of human face avatars are trained in movies. We assume that we have external tools for segmenting frames in the movie. We often use additional information like each frame's camera angle or FLAME representation. 

In~\cite{gafni2021dynamic}, authors implicitly model the facial expressions by conditioning the NeRF with the global expression code obtained from 3DMM tracking~\cite{thies2016face2face}. In \cite{zielonka2022instant}, authors leverage the idea of dynamic neural radiance fields to improve the mouth region’s rendering, which is not represented by the face model motion prior.
The IMAvatar~\cite{zheng2022avatar} model learns the subject-specific implicit representation of texture together with expression.
In \cite{gao2022reconstructing} authors use neural graphics primitives, where for each of the blend shapes, a multi-resolution grid is trained.
In RigNeRF~\cite{athar2022rignerf} authors propose a model that changes head pose and facial expressions using a deformation field that is guided
by a 3D morphable face model (3DMM).

Diverging from the previously mentioned approach, certain researchers adopt an explicit apprehension strategy. In the case of \cite{aneja2022clipface}, the authors introduced ClipFace, a method facilitating text-guided editing of textured 3D face models. In \cite{khakhulin2022realistic}, a one-shot mesh-based model reconstruction is presented, while in \cite{zielonka2022towards,danvevcek2022emoca}, a model is proposed that draws upon a blend of 2D and 3D datasets.

Our method is situated between the above approaches. Similarly to the explicit approach, we use a single 3D object instead of movies to train. We also use FLAME mesh to edit the avatar's shape end expressions. On the other hand, we use an implicit representation of the colors of objects.











\section{\our{}:  FLAME-based conditioning of NeRF for 3D face renderin}
\label{sec:method}

In this subsection, we introduce \our{} - the novel 3D face representation that combines the benefits of Flame and NeRF models. We first provide the details about the FLAME and NeRF approaches and further describe the concept of \our{} and how it can be used to control face mesh. 


\begin{figure*}[h]
    \begin{tabular}{@{}cc@{}c@{}c@{}c@{}c@{}c@{}}
	Original wives & \multicolumn{5}{c}{ New renders obtain by \our{} } \\
\includegraphics[width=0.16\textwidth, trim=100 100 100 100, clip]{img/FACE/r_5.png} &
    \includegraphics[width=0.16\textwidth, trim=50 50 50 50, clip]{img/FACE_4/rec/000.png} &
    \includegraphics[width=0.16\textwidth, trim=50 50 50 50, clip]{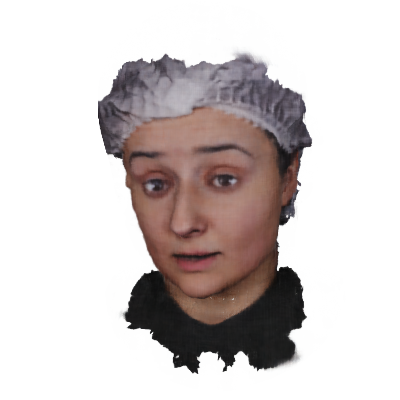} &
    \includegraphics[width=0.16\textwidth, trim=50 50 50 50, clip]{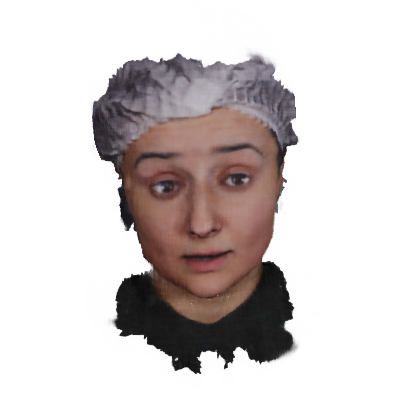} & 
    \includegraphics[width=0.16\textwidth, trim=50 50 50 50, clip]{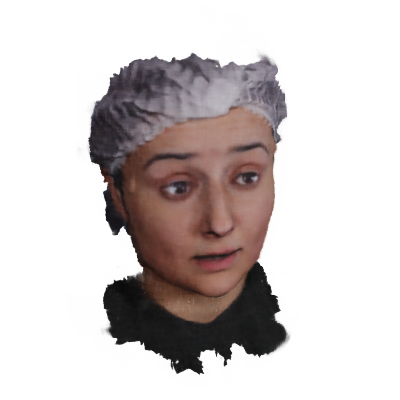} & 
    \includegraphics[width=0.16\textwidth, trim=50 50 50 50, clip]{img/FACE_4/rec/005.png}\\
    \includegraphics[width=0.16\textwidth, trim=100 100 100 100, clip]{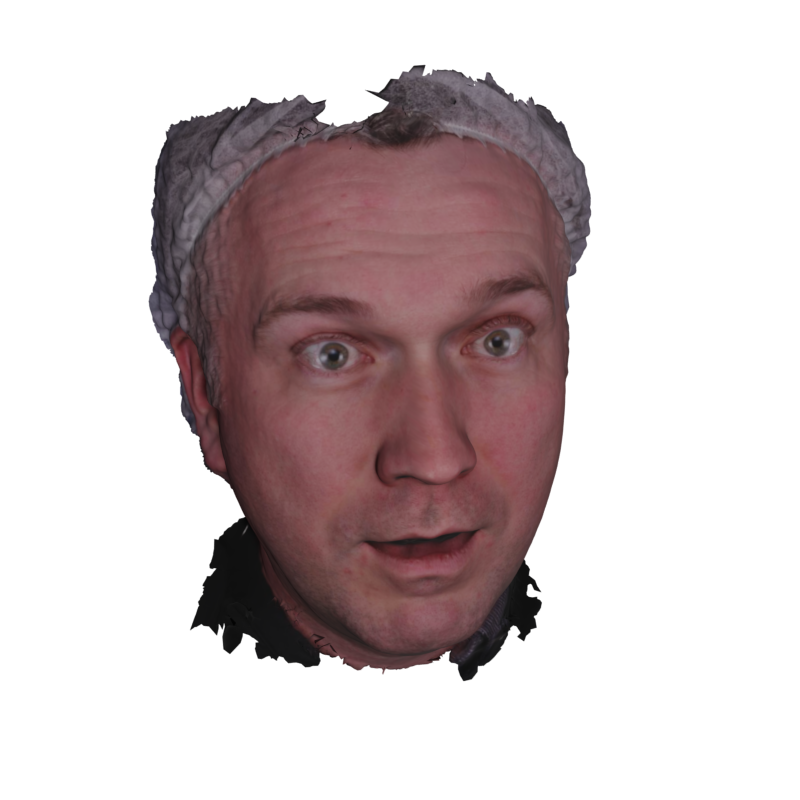} &
    \includegraphics[width=0.16\textwidth, trim=50 50 50 50, clip]{img/FACE_2/rec/000.png} &
    \includegraphics[width=0.16\textwidth, trim=50 50 50 50, clip]{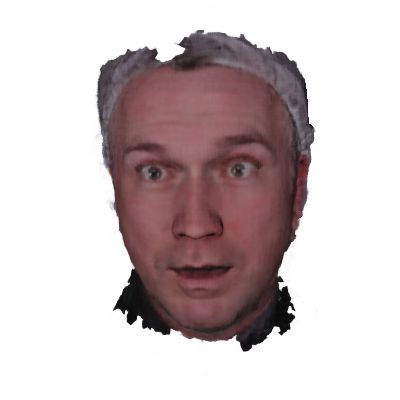} &
    \includegraphics[width=0.16\textwidth, trim=50 50 50 50, clip]{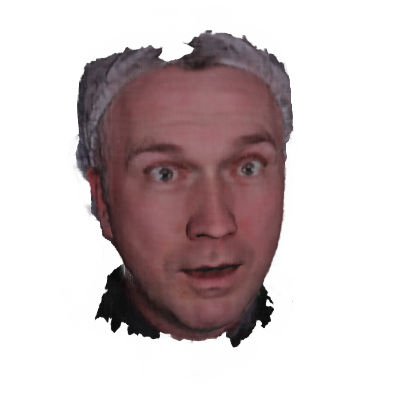} & 
    \includegraphics[width=0.16\textwidth, trim=50 50 50 50, clip]{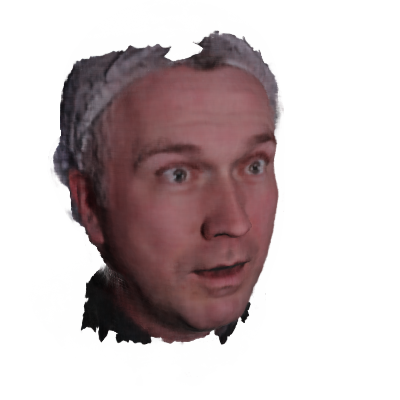} & 
    \includegraphics[width=0.16\textwidth, trim=50 50 50 50, clip]{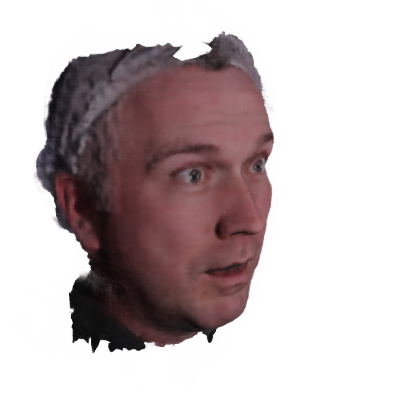}\\
    \includegraphics[width=0.16\textwidth, trim=100 100 100 100, clip]{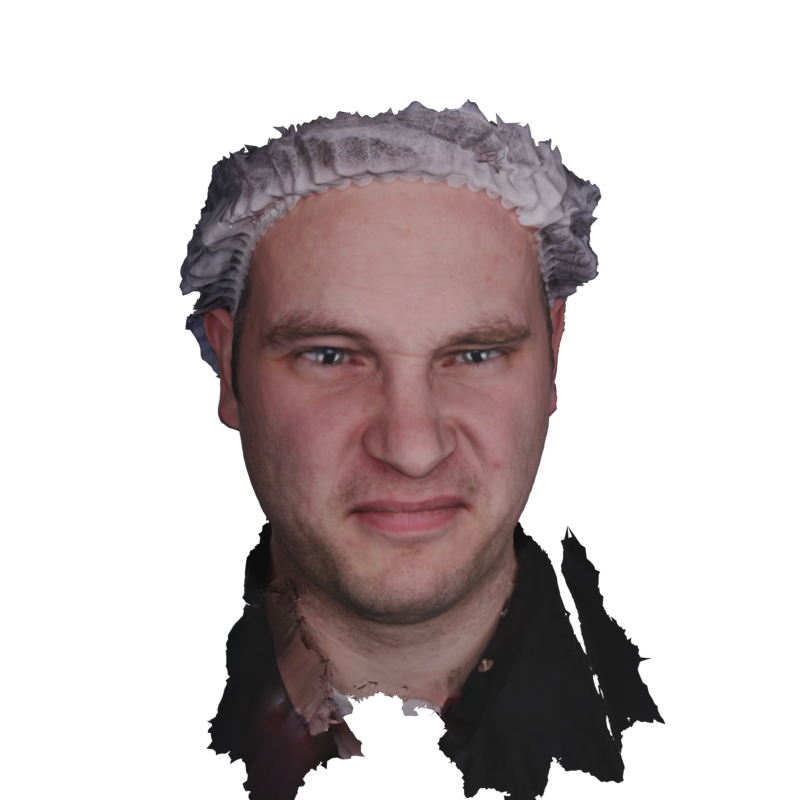} &
    \includegraphics[width=0.16\textwidth, trim=50 50 50 50, clip]{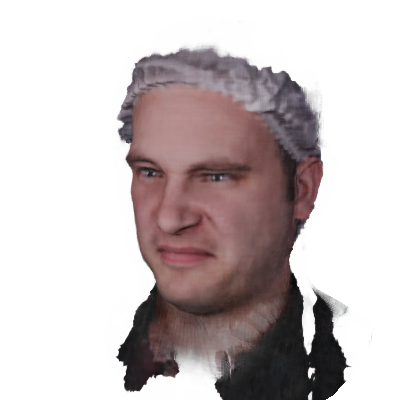} &
    \includegraphics[width=0.16\textwidth, trim=50 50 50 50, clip]{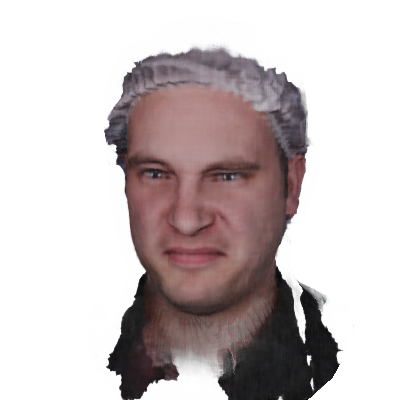} &
    \includegraphics[width=0.16\textwidth, trim=50 50 50 50, clip]{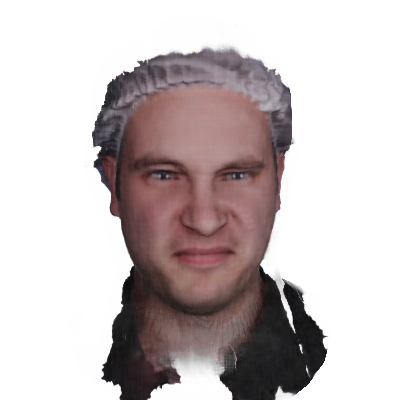} & 
    \includegraphics[width=0.16\textwidth, trim=50 50 50 50, clip]{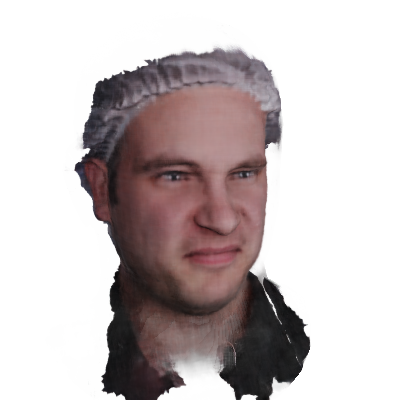} & 
    \includegraphics[width=0.16\textwidth, trim=50 50 50 50, clip]{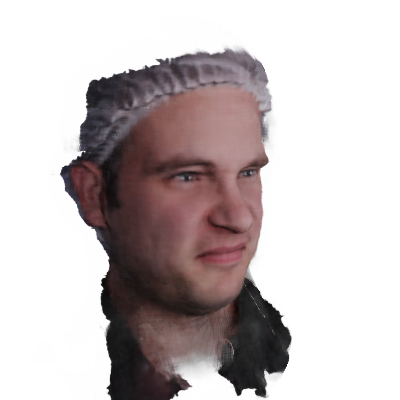}\\

\includegraphics[width=0.16\textwidth, trim=100 100 100 100, clip]{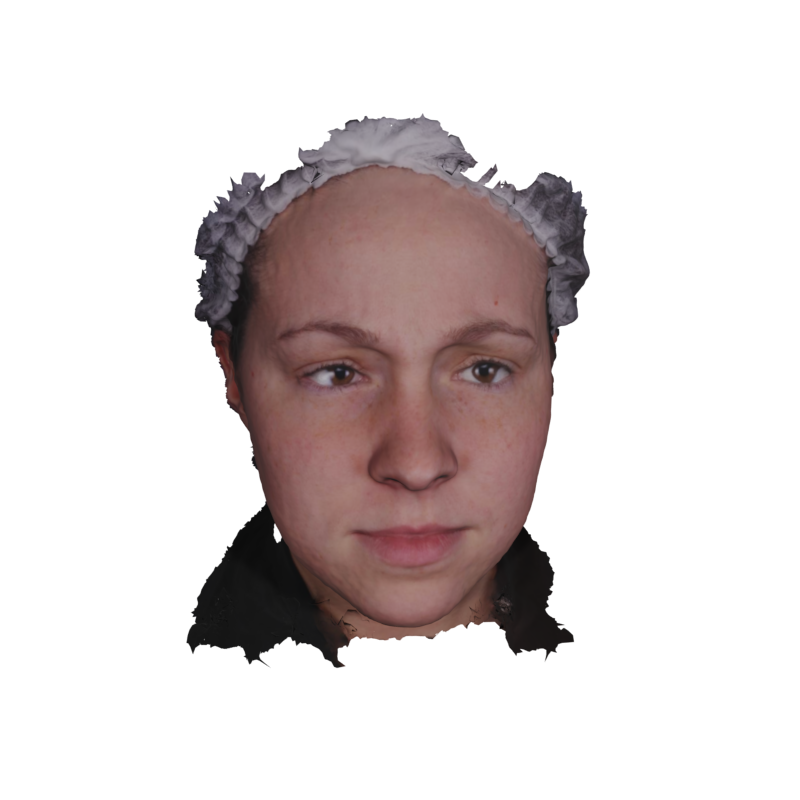} &
    \includegraphics[width=0.16\textwidth, trim=50 50 50 50, clip]{img/FACE_5/rec/000.png} &
    \includegraphics[width=0.16\textwidth, trim=50 50 50 50, clip]{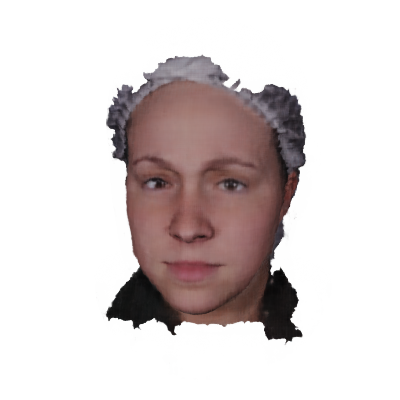} &
    \includegraphics[width=0.16\textwidth, trim=50 50 50 50, clip]{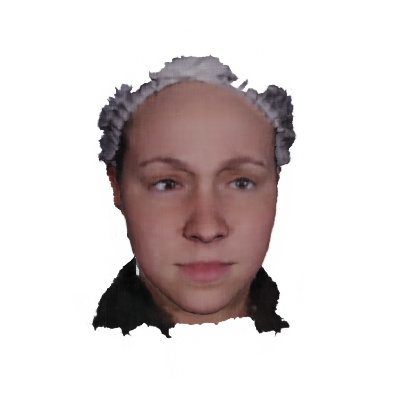} & 
    \includegraphics[width=0.16\textwidth, trim=50 50 50 50, clip]{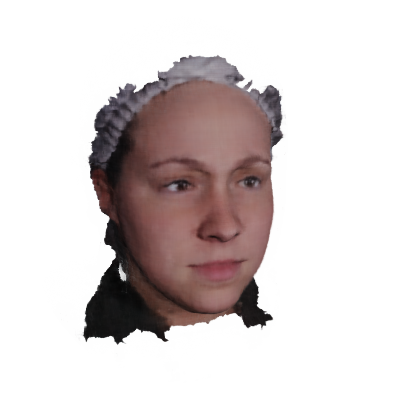} & 
    \includegraphics[width=0.16\textwidth, trim=50 50 50 50, clip]{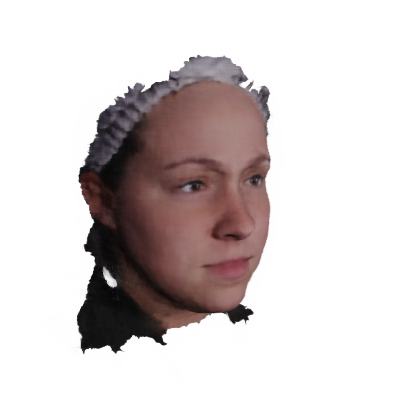}\\
    
	\end{tabular}
\caption{Reconstruction of 3D object obtained by \our{}. As we can see, \our{} model the detailed appearance of the 3D face.}
\label{fig:reconst} 
\end{figure*}

\subsection{FLAME}

FLAME (Faces Learned with an Articulated Model and Expressions)~\cite{li2017learning} is a 3D facial model trained from thousands of accurately aligned 3D scans. The model is factored in that it separates the representation of identity, pose, and facial expression, similar to the human body approach. It is represented by low polygon count, articulation, and blend skinning that is computationally efficient, compatible with existing game and rendering engines, and simple in order to maintain its practicality. The parameters of the model are trained by optimizing the reconstruction loss, assuming a detailed
temporal registration of our template mesh with three unconnected components, \mbox{including the base face and two eyeballs}. 

Formally, the FLAME is a function from human face parametrization $\F_{flame}(\beta, \psi, \phi)$ where $\beta, \psi$ and $\phi$ describe shape, expression, and pose parameters to a mesh with $n$ vertices:

\begin{equation}
\F_{flame}( \beta, \psi, \phi): \R^{k_{\beta} \times k_{\psi} \times k_{\phi}} \rightarrow \R^{n \times 3},
\end{equation}
where $k_{\beta}$, $k_{\psi}$, and $k_{\phi}$ are the numbers of shape, expression, and pose parameters.





In the classical version, we can fit our model to 3D scans or 2D images by using facial landmarks. Many strategies exist to choose landmarks and parameters for its training~\cite{li2017learning}. 
However, in the high level of generalization for input $I$ -- image 3D scan (or 2D image) and arbitrarily chosen method for estimation facial landmark points $\LP$ we minimize L2 distance 
$$
\min_{(\beta, \psi, \phi)} \| \LP(\F_{flame}( \beta, \psi, \phi)) - \LP(I) \|_2
$$
Such an approach is effective, but there are a few limitations. Localizing landmarks and choosing which parameters to optimize first is not trivial. On the other hand, for 2D images, the results are not well-qualified. We can use a pre-trained auto-encoder-based model DECA~\cite{feng2021learning} for face reconstruction from 2D images to solve such problems.

In this paper, we train the FLAME-based model in a NeRF-based scenario. As input, we take a few 2D images. As an effect, we obtain a correctly fitted FLAME model and NeRF rendering model for new views.

\subsection{NeRF}

\paragraph{NeRF representation of 3D objects}

NeRFs~\cite{mildenhall2020nerf}  is the model for representing complex 3D scenes using neural architectures. In order to do that, NeRFs take a 5D coordinate as input, which includes the spatial location $ \x = (x, y, z)$ and viewing direction ${\bf d} = (\theta, \Psi)$ and returns emitted color ${\bf c} = (r, g, b)$ and volume density $\sigma$.

A classical NeRF uses a set of images for training. In such a scenario, we produce many rays traversing through the image and a 3D object represented by a neural network. 
NeRF parameterized by $\bf \Theta$ approximates this 3D object with an MLP network:
$$
\F_{NeRF} (\x , \bf d; \Theta ) = ( {\bf c} , \sigma).
$$
The model is trained to map each input 5D coordinate to its corresponding volume density and directional emitted color.  

The loss of NeRF is inspired by classical volume rendering \cite{kajiya1984ray}.  We render the color of all rays passing through the scene. The volume density $\sigma( \x )$ can be interpreted as the differential probability of a ray. The expected color
$C({\bf r})$ of camera ray ${\bf r}(t) = {\bf o} + t {\textit{d}}$ (where  ${\bf o}$ is ray origin and ${\textit{d}}$ is direction) can be computed with an integral.

In practice, this continuous integral is numerically estimated using a quadrature. We use a stratified sampling approach where we partition our ray $[t_n, t_f ]$ into $N$ evenly-spaced bins and then draw one sample $t_i$ uniformly at random from within each bin. We use these samples to~estimate~$C({\bf r})$ with the quadrature rule~\cite{max1995optical}, where $\delta_i = t_{i+1} - t_i$ is the distance between adjacent samples:

\begin{equation*}
\hat C({\bf r}) = \sum_{i=1}^{N} T_i (1-\exp(-\sigma_i \delta_i)) {\bf c}_i, 
\end{equation*}
$$
\mbox{ where } T_i=\exp \left(-\sum_{j=1}^{i-1} \sigma_j \delta_j \right),
$$
This function for calculating $\hat C({\bf r})$ from the set of $(\bf c_i , \sigma_i)$ values is trivially differentiable.

We then use the volume rendering procedure to render the color of each ray
from both sets of samples. Contrary to the baseline NeRF~\cite{mildenhall2020nerf}, where two "coarse" and "fine" models were simultaneously trained, we use only the "coarse" architecture.
Our loss is simply the total squared error between the rendered and true pixel colors 
\begin{equation}
    \L = \sum _{{\bf r} \in R} \| \hat C({\bf r}) - C({\bf r}) \|_2^2
    \label{eq:cost_general}
\end{equation}

where $R$ is the set of rays in each batch, and $C({\bf r})$, $\hat C ({\bf r})$ are the ground truth and predicted RGB colors for ray {\bf r } respectively. 

\begin{figure}[h]
    \begin{tabular}{@{}cc@{}c@{}c@{}c@{}c@{}c@{}}
	Original wives & \multicolumn{2}{c}{ Mesh fitted by \our{} } \\
    \includegraphics[width=0.16\textwidth, trim=25 25 25 25, clip]{img/FACE_2/rec/003.png} &
    \includegraphics[width=0.16\textwidth, trim=50 50 50 50, clip]{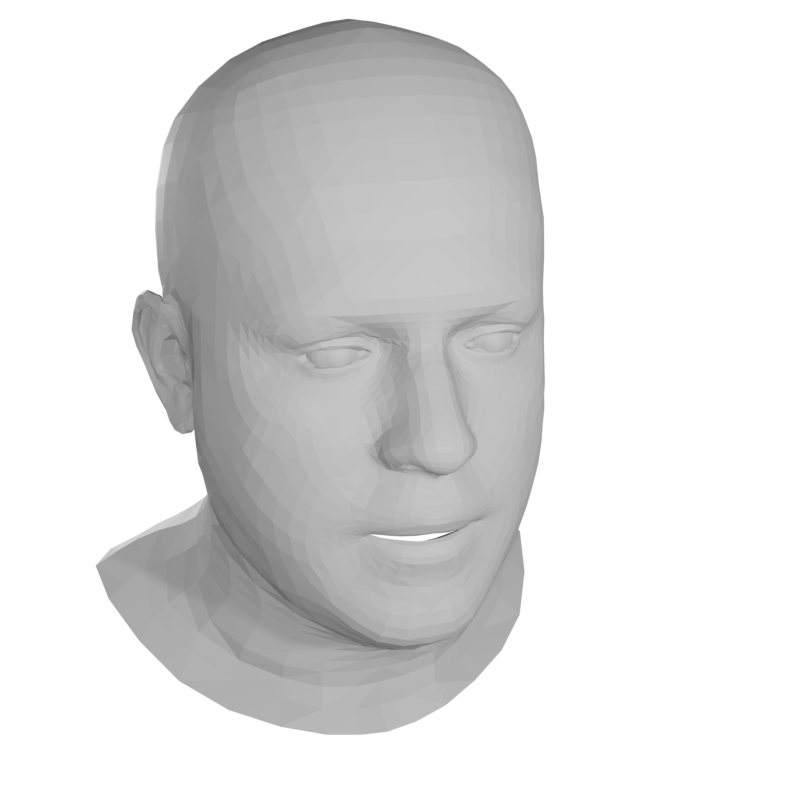} & 
    \includegraphics[width=0.16\textwidth, trim=50 50 50 50, clip]{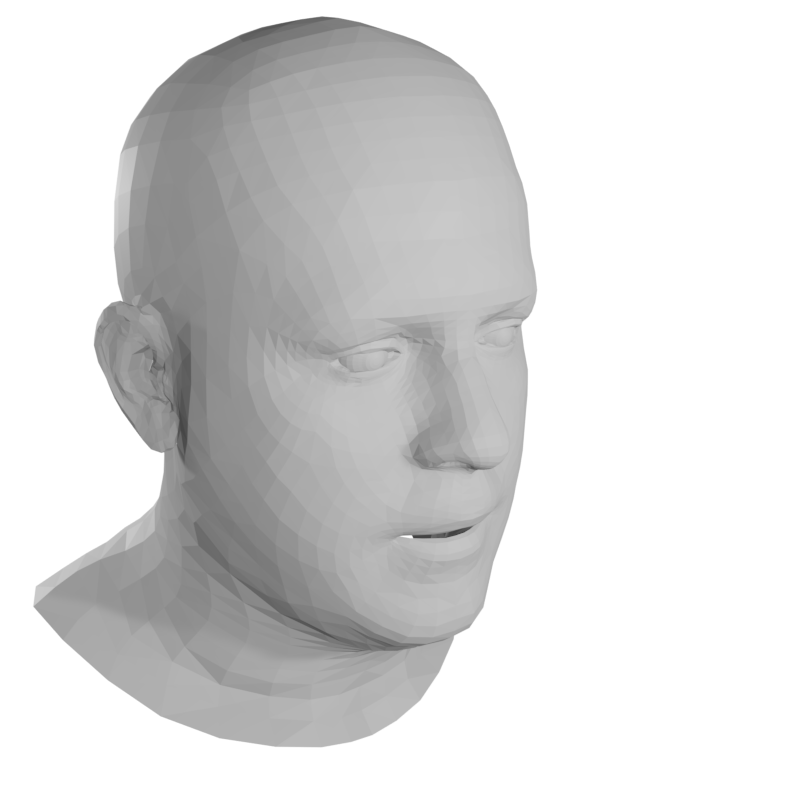}\\
    \includegraphics[width=0.16\textwidth, trim=25 25 25 25, clip]{img/FACE_3/rec/003.png} &
    \includegraphics[width=0.16\textwidth, trim=50 50 50 50, clip]{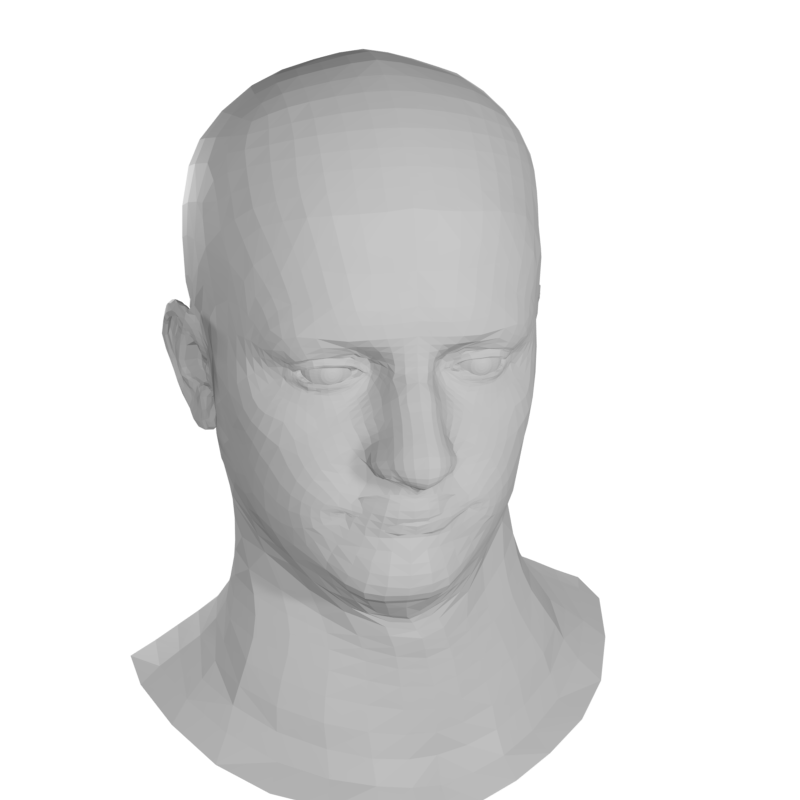} & 
    \includegraphics[width=0.16\textwidth, trim=50 50 50 50, clip]{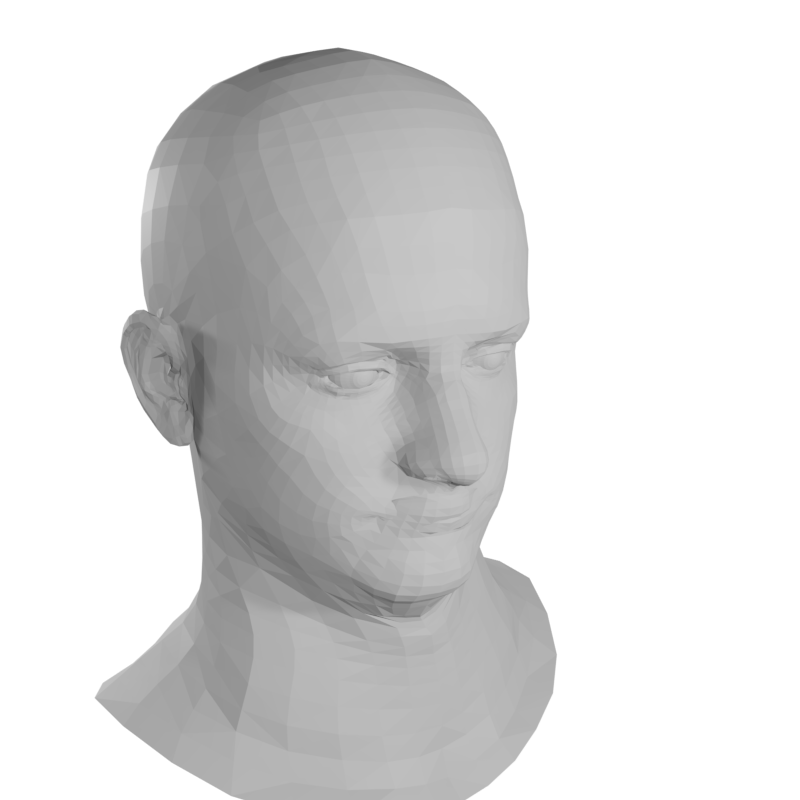}\\
    \includegraphics[width=0.16\textwidth, trim=25 25 25 25, clip]{img/FACE_5/rec/003.png} &
    \includegraphics[width=0.16\textwidth, trim=50 50 50 50, clip]{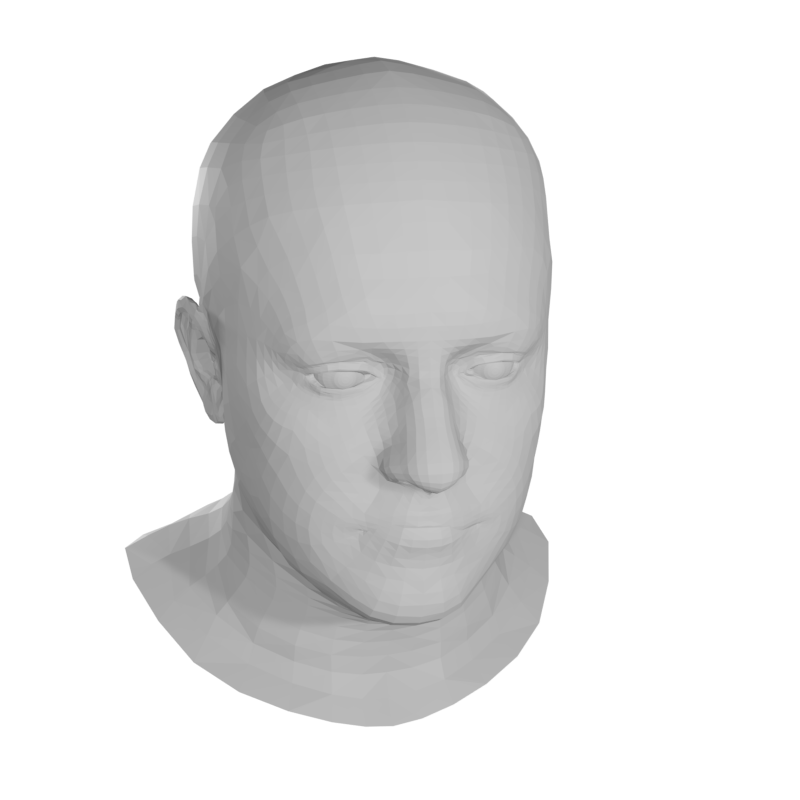} & 
    \includegraphics[width=0.16\textwidth, trim=50 50 50 50, clip]{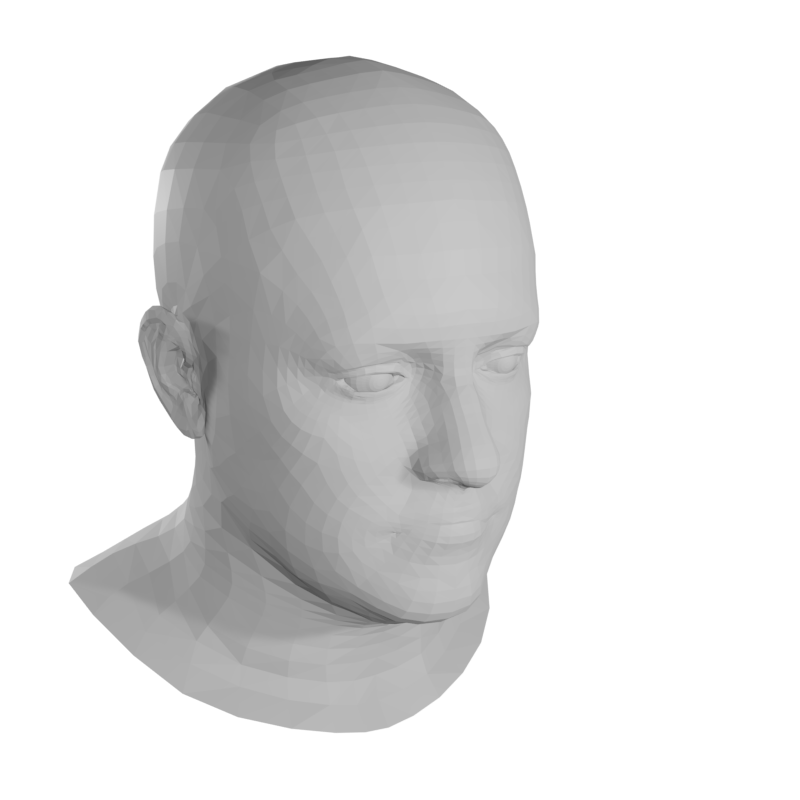}\\
    
	\end{tabular}
\caption{During the training of \our, we simultaneously model FLAME mesh and NeRf dedicated to colors. In the above figure, we present the meshes fitted by \our{}.}
\label{fig:mesh} 
\end{figure}

\subsection{\our{}}

We introduce \our{} the 3D face model that combines the benefits of mesh representations from FLAME and NeRF implicit representation of 3D objects. Thanks to the application of NeRFs, we can estimate the parameters of FLAME directly from 2D images without using landmarks points. On the other hand, we obtain NeRF model, which can be editable similarly to FLAME. In order to achieve that, we introduce the function that approximates the volume density using the FLAME model. 

Consider the distance function $d(\x,\mathcal{M})$ between point $\x = (x,y,z) \in \R^3$ and the mesh $\mathcal{M}:=\mathcal{M}_{\beta, \psi, \phi}$ created by $F_{flame}(\beta, \psi, \phi)$, with parameters $\beta, \psi, \phi$. Note, that edges between vertices in FLAME model can be directly taken from the template mesh (see \cite{athar2021flame} for details). We define the volume density function as:

\begin{equation}
   \sigma(\x,\mathcal{M})  \!\! =\!\!  \left\{
  \begin{array}{@{}ll@{}}
    0, & \text{if}\  d(\x,\mathcal{M})> \varepsilon\\
    (1 - \frac{1}{\varepsilon} d(\x,\mathcal{M})), & \text{otherwise},
  \end{array}\right.
  \label{eq:dist}
\end{equation}

where $\varepsilon$ is a hyperparameter that defines the neighborhood of the mesh surface. In practice, the values of the density volume function are non-zero only in the close neighborhood of the mesh. 


The \our{} can be represented by the function:

\begin{equation}
\F_{ \our{} }( \x ; \beta, \psi, \phi, \Theta) = ( \F^{c}_{\Theta}( \x ) , \sigma(\x,\mathcal{M} )),
\label{eq:our}
\end{equation}
where $\F^{c}_{\Theta}$ is the MLP that predicts the color, similar to the NeRF model.  


The model is trained in an end-to-end manner directly, optimizing the criterion \eqref{eq:cost_general} with respect to the MLP parameters~$\Theta$, and FLAME parameters $\beta, \psi$, $\phi$, which describe shape, expression, and pose. In \our{}, we utilize the original loss function used to train NeRF models. Therefore, the structure of colors on rays must be consistent. During training, the model modified the mesh structure to be consistent with the 3D structure of the target face. The simultaneous neural network $\F$ produces colors for the rendering procedure.

\begin{figure*}[h]
    \begin{tabular}{@{}cc@{}c@{}c@{}c@{}c@{}c@{}}
	Original position & \multicolumn{5}{c}{ Yawning } \\
    \multirow{2}{*}{
    \includegraphics[width=0.15\textwidth, trim=50 50 50 50, clip]{img/FACE_2/rec/003.png}
    } 
    &
    \includegraphics[width=0.15\textwidth, trim=50 50 50 50, clip]{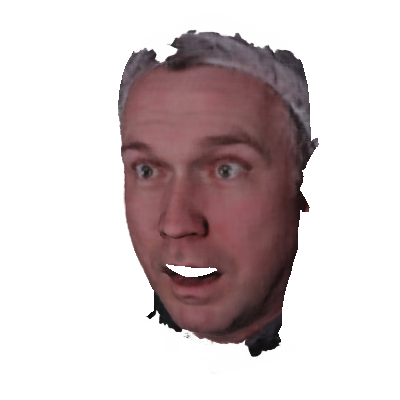} &
    \includegraphics[width=0.15\textwidth, trim=50 50 50 50, clip]{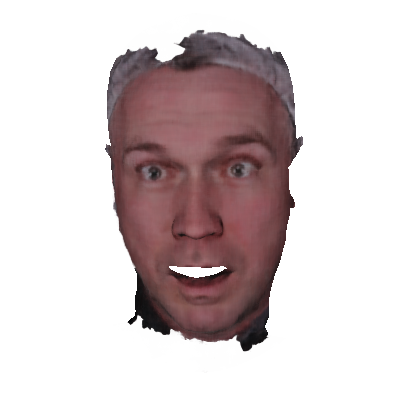} &
    \includegraphics[width=0.15\textwidth, trim=50 50 50 50, clip]{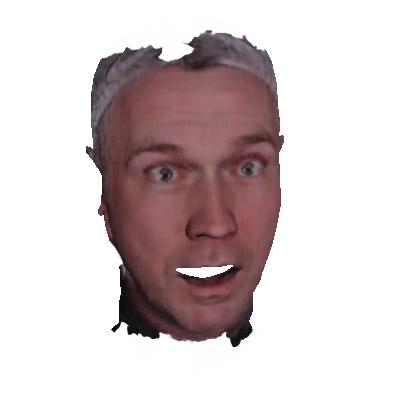} & 
    \includegraphics[width=0.15\textwidth, trim=50 50 50 50, clip]{img/FACE_2/mouth/004.png} & 
    \includegraphics[width=0.15\textwidth, trim=50 50 50 50, clip]{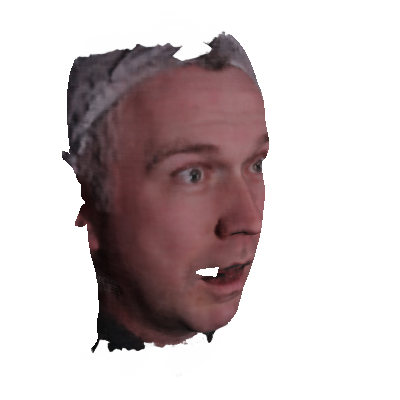}\\ 
	 & \multicolumn{5}{c}{ Smiling} \\
    &
    \includegraphics[width=0.15\textwidth, trim=50 50 50 50, clip]{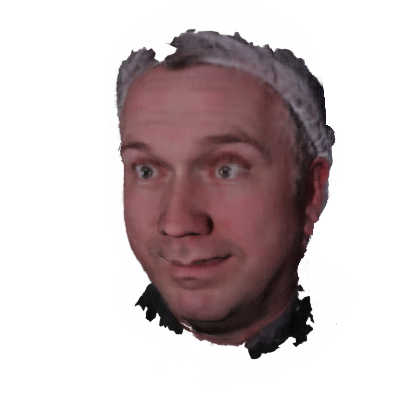} &
    \includegraphics[width=0.15\textwidth, trim=50 50 50 50, clip]{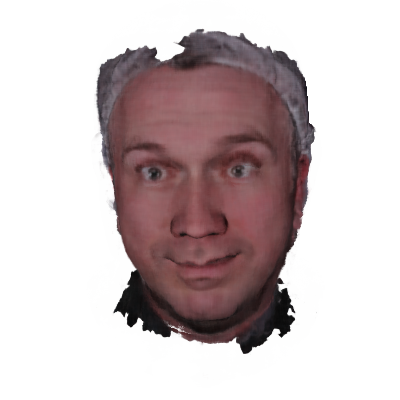} &
    \includegraphics[width=0.15\textwidth, trim=50 50 50 50, clip]{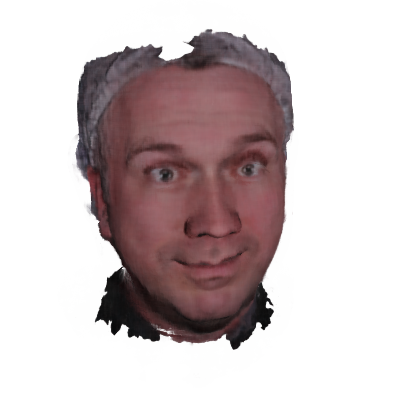} & 
    \includegraphics[width=0.15\textwidth, trim=50 50 50 50, clip]{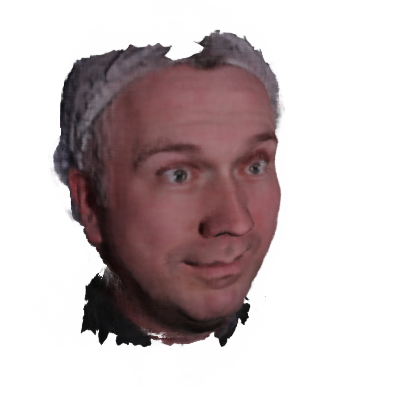} & 
    \includegraphics[width=0.15\textwidth, trim=50 50 50 50, clip]{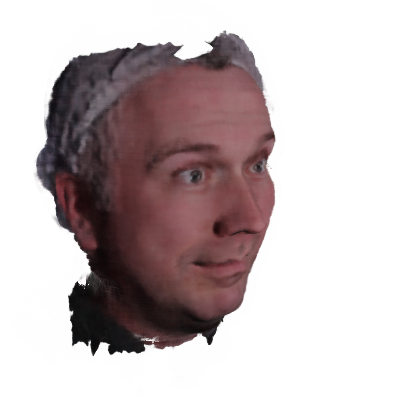}\\
	Original position & \multicolumn{5}{c}{ Yawning } \\
    \multirow{2}{*}{
    \includegraphics[width=0.15\textwidth, trim=50 50 50 50, clip]{img/FACE_5/rec/003.png}
    } 
    &
    \includegraphics[width=0.15\textwidth, trim=50 50 50 50, clip]{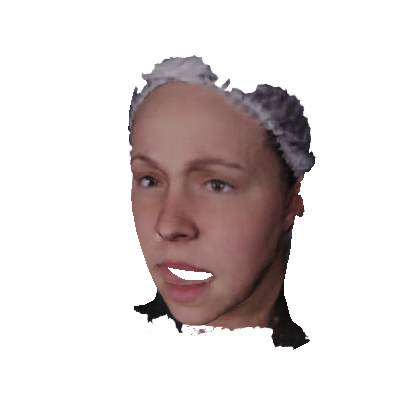} &
    \includegraphics[width=0.15\textwidth, trim=50 50 50 50, clip]{img/FACE_5/mouth/002.png} &
    \includegraphics[width=0.15\textwidth, trim=50 50 50 50, clip]{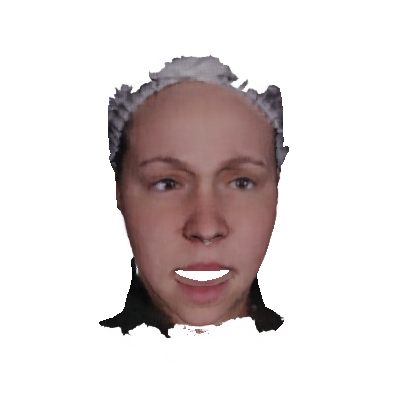} & 
    \includegraphics[width=0.15\textwidth, trim=50 50 50 50, clip]{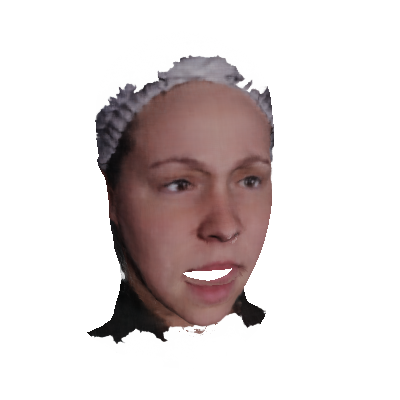} & 
    \includegraphics[width=0.15\textwidth, trim=50 50 50 50, clip]{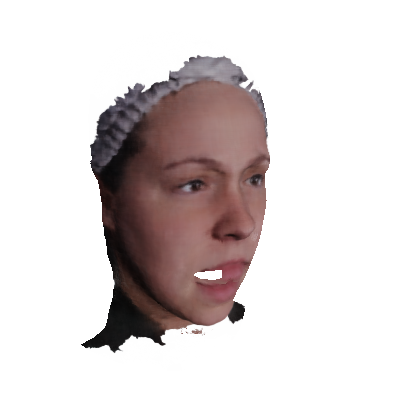}\\ 
	 & \multicolumn{5}{c}{ Smiling} \\
    &
    \includegraphics[width=0.15\textwidth, trim=50 50 50 50, clip]{img/FACE_5/rot/000.png} &
    \includegraphics[width=0.15\textwidth, trim=50 50 50 50, clip]{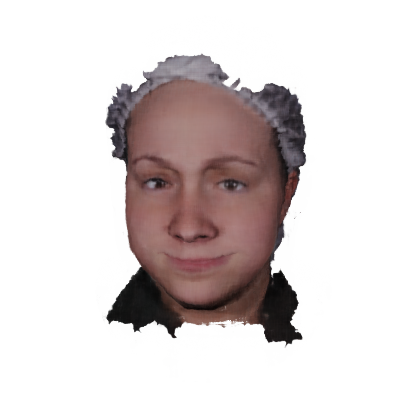} &
    \includegraphics[width=0.15\textwidth, trim=50 50 50 50, clip]{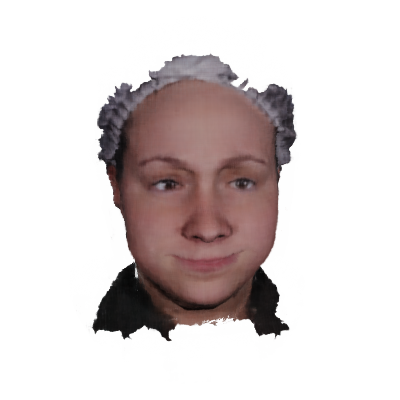} & 
    \includegraphics[width=0.15\textwidth, trim=50 50 50 50, clip]{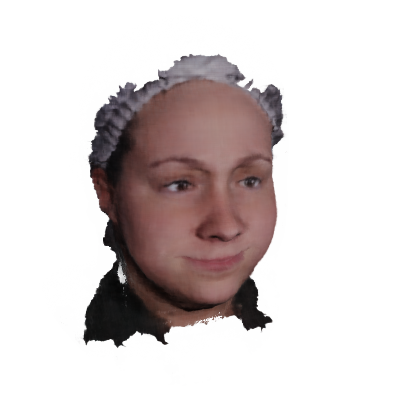} & 
    \includegraphics[width=0.15\textwidth, trim=50 50 50 50, clip]{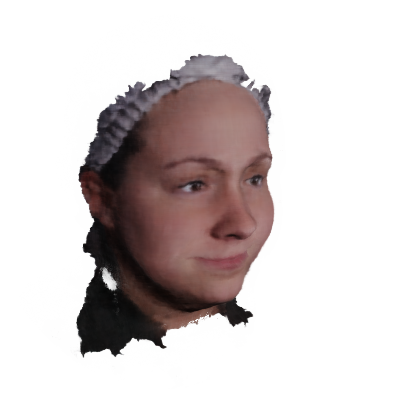}\\
	\end{tabular}
\caption{Our model allows producing manipulation of the human face. In \our{}, we use FLAME for volume density rendering. Therefore, we can manipulate FLAME features and modify NeRF representation. In the figure, we show three faces and their versions with open mouths and changing expressions from different views.  }
\vspace{1cm}
\label{fig:FLAME_compare1} 
\end{figure*}

During training, we can see the trade-off between the level of mesh fitting and the quality of renders.
The main reason is that we must train our model with small $\varepsilon$, and color must be encoded in a small neighborhood of the mesh. Therefore, the quality of the model is lower than classical NeRF. 
With larger $\varepsilon$ we obtain mesh which is not correctly fitted. 

Therefore, we use the MLP $\F^{\sigma}_{\Theta}$  that predicts the volume density analogical to NeRF. In the first 10 000 epoch we train our model with volume density given by formula (\ref{eq:dist}). Then, Flame parameters are frozen, and we train only NeRF component 
with volume density given by 
\begin{equation*}
   \sigma(\x,\mathcal{M}, \F^{\sigma}_{\Theta})  \!\! =\!\!  \left\{
  \begin{array}{@{}ll@{}}
    0, & \text{if}\  d(\x,\mathcal{M})> \varepsilon\\
    \F^{\sigma}_{\Theta}(\x), & \text{otherwise},
  \end{array}\right.
\end{equation*} 

where $\varepsilon$ is a hyperparameter that defines the neighborhood of the mesh surface and $F^{\sigma}_{\Theta}$ is MLP that predicts volume density. The value of $\varepsilon$ increases gradually over time during training after 10 000 epochs. The value of $\varepsilon$ increases up to $\varepsilon=0.1$ at the end of the training procedure.

\subsection{Controlling NeRF models to obtain face manipulation}

The classical NeRF method is known to generate highly detailed and realistic images. However, it can be challenging to manipulate NeRF models to achieve precise facial modifications. Several techniques, such as generative models, dynamic scene encoding, and conditioning mechanisms, have been proposed to address this challenge. Nonetheless, controlling NeRF models to the same extent as mesh representations remains elusive. In contrast, the FLAME is a straightforward model with three parameters, namely $\beta$, $\psi$, and $\phi$, representing shape, expression, and pose, respectively. By performing simple linear operations on these parameters, it is possible to rotate the avatar, change facial expressions, and adjust facial features to a certain degree.

\begin{figure}[!h]
    \begin{center}
    \usetikzlibrary{shapes.geometric,positioning}
    \begin{tikzpicture}[
      triangle/.style={
       regular polygon,
       regular polygon sides=3,
       minimum size=2cm,
       draw,
       }
    ]
    \node [triangle] (a) {};
    \node [triangle,rotate=40,right=3cm of a] (b) {};
    \draw [<-,shorten <=4mm,shorten >=4mm] (a) to[bend left] node[above]{$T( \cdot )$} (b);
        \node[text width=3.5cm] at (0.5,-0.5) {$ p_3 $};
        \node[text width=3.5cm] at (2.7,-0.5) {$ p_1 $};
        \node[text width=3.5cm] at (1.6,1.3) {$ p_2 $};

        \node[text width=3.5cm] at (5.3,-0.8) {$ q_1 $};
        \node[text width=3.5cm] at (4.8,1.3) {$ q_3 $};
        \node[text width=3.5cm] at (6.9,0.8) {$ q_2 $};

        \draw[red, thick] (4.2,0.3) -- (4.8,2.3);
        \draw[red, thick, dashed] (3.6,-1.7) -- (4.2,0.3);

        \draw[red, thick, dashed, rotate=-40] (0.0,0.0) -- (0.6,1.9);
        \draw[red, thick, rotate=-40] (-0.6,-1.8) -- (0.0,0.0);
        
    \end{tikzpicture}
    
    \caption{Visualization of transformation in \our{}. We aim to aggregate colors along the ray during rendering in the new position (see the red line in the right image). \our{} uses FLAME mesh, therefore we can localize the face's vertex, which is crossed with the ray $p_1, p_2,p_3$ and the corresponding triangle in the initial position mesh $q_1, q_2,q_3$. Thanks to such pairs of points, we estimate affine transformation $T$, which is used to find the ray in the initial position (see the red line in the left image). } 
    \label{fig:transformation}
    \end{center}
    \vspace{-0.5cm}
\end{figure}
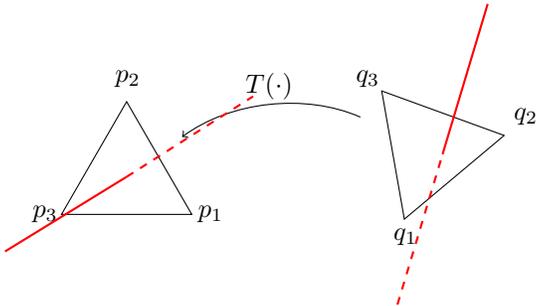

Our \our{} is built on the FLAME model so that we can manipulate our FLAME model to control density prediction $\sigma$. However, the problem is with predicting RGB colors after transformation. To solve such a problem, we use a simple technique. For the prediction of color after transformation, we will return to the starting position where the color is known, see Fig.~\ref{fig:transformation}.  

Let us consider \our{} model, which is already trained. We have parameters $\beta_1, \psi_1, \phi_1, \Theta, \varepsilon$, function
$$
\F_{ \our{} }( \x ; \beta_1, \psi_1, \phi_1, \Theta) 
$$
and fitted mesh $\mathcal{M}_1$, created by the FLAME from parameters $\beta_1, \psi_1, \phi_1$. Let us assume that we apply some modification of FLAME parameters, which means that we obtain new $\beta_2, \psi_2, \phi_2$. Using these parameters, we can create the modified mesh $\mathcal{M}_2$, simply using the FLAME model. Instead of retraining the NeRF model for new parameters requiring the 2D images for a new pose, we propose applying a simple transformation between the modified and original space.

We consider some point $\x_2$ located on mesh $\mathcal{M}_2$ representing the new pose. We postulate using the affine transformation $T(\cdot)$, which transforms the point on the mesh to the original pose, in which the model was trained:
$
    T(\x_2) = \x_1,
$
where $\x_1 \in \mathcal{M}_1$ is the corresponding point on original pose. After the transformation, we can identify an element in the original pose for each element on the mesh. In practice, finding the transformation $T$ is not trivial since it depends on the local transformation of the mesh. 

However, for a given point $\x_1 \in \mathcal{M}_1$, we can find a face triangle described by vertices $(\q_1,\q_2,\q_3) \in V_1$ that contains $\x_1$, where $V_1$ is the set of vertices for mesh $\mathcal{M}_1$. Because FLAME model is shifting the vertices of the model keeping the connections unchanged, we can locate the corresponding triangle $(\p_1,\p_2,\p_3) \in V_2$ in mesh  $\mathcal{M}_2$. For each of the triangles that create the mesh, we define affine transformation:

\begin{equation}
T_m(\p_i) = \q_i, \mbox{ for } i =1,2,3.
\end{equation}


In such a situation, we assume that such transformations $T_m(\cdot)$ are affine, and we can use Least-Squares Conformal Multilinear Regression~\cite{schmid2012tridimensional} to estimate the parameters. Practically, finding the transformation for each of the triangles is extremely fast, requires inverting a fourth-dimensional matrix, and can be parallelized. Having the parameters of transformations estimated, we can apply them directly to the $\x$ in \our{} model given by \eqref{eq:our} and calculate the colors as in an unshifted pose.  





\begin{table*}
\vspace{0.5cm}
\begin{center}
    \begin{tabular}{cccc|ccc|ccc}
     & \multicolumn{3}{c}{ PSNR $\uparrow$ }  & \multicolumn{3}{c}{ SSIM $\uparrow$ } & \multicolumn{3}{c}{ LPIPS $\downarrow$ } \\[0.1ex]
    & NeRF &  \our{} & FLAME  & NeRF &  \our{} & FLAME & NeRF &  \our{} & FLAME\\[0.1ex] 
     \hline
     Face 1 &  33.37 & 27.89 &  9.67 &  0.96 & 0.95 & 0.76 &  0.05 &  0.09 & 0.26 \\[0.1ex]
     Face 2 &  33.39 & 29.79 &  12.44 &  0.96 & 0.96 & 0.80 &  0.05 &  0.06 & 0.24 \\[0.1ex] 
     Face 3 &  33.08 & 29.70 &  12.97 &  0.97 & 0.95 & 0.82 &  0.04 &  0.08 & 0.20 \\[0.1ex]
     Face 4 &  31.96 & 25.78 &  12.51 &  0.96 & 0.92 & 0.79 &  0.04 &  0.10 & 0.23 \\[0.1ex] 
     Face 5 &  33.15 & 32.59 &  11.30 &  0.96 & 0.96 & 0.77 &  0.05 &  0.05 & 0.26 \\[0.1ex] 
    Face 6 &  32.42 & 29.18 &  11.45 &  0.96 & 0.95 & 0.76 &  0.06 &  0.07 & 0.26 \\[0.1ex]
    \end{tabular}
\caption{Comparison of PSNR, SSIM, and LPIPS matrices between our model and NeRF and FLAME baselines. As we can see NeRF gives essentially better since do not allow manipulation. On the other hand, FLAME model gives inferior results since we train mash and texture only on landmark points. }
\label{tab:nerf}
\end{center}
\end{table*}



Our approach is sensitive to some particular facial manipulations. When we open the mouth of our avatar, we obtain artifacts. Three main reasons cause such problems. First, it is difficult to fit the mouth around the mesh to images since it is very sensitive to perturbations.
Additionally, the mesh lacks internal content, and it cannot represent the inside of the mouth and tongue. The third problem is that rays go through the open mouth, cut the mesh back of the head, and render some artifacts.

To solve such a problem, we remove rays through an open-mouth region. Such a solution is simple to implement since we can easily filter rays that do not cross mesh. On the other hand, allows for reducing most of the artifacts.

\section{Experiments}

In this section, we describe the experimental results of the proposed model. To our knowledge, it is the first model that obtains editable NeRF trains on a single object in one position. Most of the methods use movies to encode many different positions of the face.  
We can produce novel views in training positions and in modified facial expressions using knowledge only from one fixed position. 
Therefore, it is hard to compare our results to other algorithms. 
In the first subsection, we show that our model produces high-quality NeRF representations of the objects by comparing our model with our baseline classical NeRF and classical textured FLAME. In the second subsection, we present meshes obtained by our model. Finally, we show that our model allows facing manipulations.

Since the current literature does not provide suitable data sets for evaluating the NeRF-based model for modeling 3D face avatars, we created a data set using 3D scenes. We create a classical NeRF training data set. We construct \mbox{$200 \times 200$} transparent background images from random positions.

\subsection{Reconstruction Quality}
In this subsection, we show that \our{} can reconstruct a 3D human face with similar quality as classical NeRF. Since we train our model on a single position, it is difficult to compare our model to dynamic neural radiance fields. Therefore we show that our model has a slightly lower quality than classical NeRF but allows dynamic modification. On the other hand, we show that we obtain better results than textured FLAME, which cannot capture the geometry and appearance details of the human face. 

In Fig~\ref{fig:FLAME_compare}, we compare \our{} and textured FLAME. As we can see, \our{} can reproduce facial features and geometry. On the other hand, FLAME produces well-suited textures, but the mesh is not well-suited. In Tab.~\ref{tab:nerf}, we present a numerical comparison.
We compare the metric reported by NeRF called PSNR (\textit{peak signal-to-noise ratio}), SSIM (\textit{structural similarity index measure}), LPIPS (\textit{learned perceptual image patch similarity}) used to measure image reconstruction effectiveness.
As we can see NeRF gives essentially better since do not allow manipulation. On the other hand, FLAME model gives inferior results since we train mash and texture only on landmark points.
In Fig.~\ref{fig:reconst}, we present new renders of the model obtained by \our{}. As we can see, \our{} model the detailed appearance of the 3D face.

\subsection{Mesh fitting}

The RGB colors generated by NeRF are present only in the $\varepsilon$ vicinity of the mesh. This approach allows for a precise fitting of the mesh to the human face, which is critical for generating animated models. In Figure~\ref{fig:mesh}, we present the rendered faces and corresponding meshes produced by our \our{} approach. The results demonstrate that our method can accurately capture the underlying mesh \mbox{structure}.

\subsection{Face manipulation}


Our approach enables the manipulation of human facial features. Leveraging FLAME as a backbone, \our{} offers the ability to manipulate FLAME features and modify NeRF representations. In Figure~\ref{fig:FLAME_compare1}, we showcase three faces and their modification including open mouths and changing expressions, that can be manipulated using our model in a manner similar to the classical FLAME model.

\our{} simultaneously train mesh and NeRF components for color. After training, we can exchange the produced mesh in our model to obtain a modification of the final look of the avatar, see Fig.~\ref{fig:FLAME_compare_tezer}.

\section{Conclusions}
\label{sec:conclusions}

In this work, we introduce a novel approach called \our{}, which combines NeRF and FLAME to achieve high-quality rendering and precise pose control. While NeRF-based models use neural networks to model RGB colors and volume density, our method utilizes an explicit density volume represented by the FLAME mesh. This allows us to model the quality of NeRF rendering and accurately control the appearance of the final output. As a result of offering complete control over the model, the quantitative performance of our approach is marginally inferior to that of a static NeRF model.


{\small

\begin{thebibliography}{10}\itemsep=-1pt

\bibitem{achlioptas2018learning}
Panos Achlioptas, Olga Diamanti, Ioannis Mitliagkas, and Leonidas Guibas.
\newblock Learning representations and generative models for 3d point clouds.
\newblock In {\em International conference on machine learning}, pages 40--49.
  PMLR, 2018.

\bibitem{aneja2022clipface}
Shivangi Aneja, Justus Thies, Angela Dai, and Matthias Nie{\ss}ner.
\newblock Clipface: Text-guided editing of textured 3d morphable models.
\newblock {\em arXiv preprint arXiv:2212.01406}, 2022.

\bibitem{arsalan2017synthesizing}
Amir Arsalan~Soltani, Haibin Huang, Jiajun Wu, Tejas~D Kulkarni, and Joshua~B
  Tenenbaum.
\newblock Synthesizing 3d shapes via modeling multi-view depth maps and
  silhouettes with deep generative networks.
\newblock In {\em Proceedings of the IEEE conference on computer vision and
  pattern recognition}, pages 1511--1519, 2017.

\bibitem{athar2021flame}
ShahRukh Athar, Zhixin Shu, and Dimitris Samaras.
\newblock Flame-in-nerf: Neural control of radiance fields for free view face
  animation.
\newblock {\em arXiv preprint arXiv:2108.04913}, 2021.

\bibitem{athar2022rignerf}
ShahRukh Athar, Zexiang Xu, Kalyan Sunkavalli, Eli Shechtman, and Zhixin Shu.
\newblock Rignerf: Fully controllable neural 3d portraits.
\newblock In {\em Proceedings of the IEEE/CVF Conference on Computer Vision and
  Pattern Recognition}, pages 20364--20373, 2022.

\bibitem{azinovic2022neural}
Dejan Azinovi{\'c}, Ricardo Martin-Brualla, Dan~B Goldman, Matthias
  Nie{\ss}ner, and Justus Thies.
\newblock Neural rgb-d surface reconstruction.
\newblock In {\em Proceedings of the IEEE/CVF Conference on Computer Vision and
  Pattern Recognition}, pages 6290--6301, 2022.

\bibitem{barron2021mip}
Jonathan~T Barron, Ben Mildenhall, Matthew Tancik, Peter Hedman, Ricardo
  Martin-Brualla, and Pratul~P Srinivasan.
\newblock Mip-nerf: A multiscale representation for anti-aliasing neural
  radiance fields.
\newblock In {\em Proceedings of the IEEE/CVF International Conference on
  Computer Vision}, pages 5855--5864, 2021.

\bibitem{barron2022mip}
Jonathan~T Barron, Ben Mildenhall, Dor Verbin, Pratul~P Srinivasan, and Peter
  Hedman.
\newblock Mip-nerf 360: Unbounded anti-aliased neural radiance fields.
\newblock In {\em Proceedings of the IEEE/CVF Conference on Computer Vision and
  Pattern Recognition}, pages 5470--5479, 2022.

\bibitem{chan2022efficient}
Eric~R Chan, Connor~Z Lin, Matthew~A Chan, Koki Nagano, Boxiao Pan, Shalini
  De~Mello, Orazio Gallo, Leonidas~J Guibas, Jonathan Tremblay, Sameh Khamis,
  et~al.
\newblock Efficient geometry-aware 3d generative adversarial networks.
\newblock In {\em Proceedings of the IEEE/CVF Conference on Computer Vision and
  Pattern Recognition}, pages 16123--16133, 2022.

\bibitem{chen2022tensorf}
Anpei Chen, Zexiang Xu, Andreas Geiger, Jingyi Yu, and Hao Su.
\newblock Tensorf: Tensorial radiance fields.
\newblock In {\em Computer Vision--ECCV 2022: 17th European Conference, Tel
  Aviv, Israel, October 23--27, 2022, Proceedings, Part XXXII}, pages 333--350.
  Springer, 2022.

\bibitem{choy20163d}
Christopher~B Choy, Danfei Xu, JunYoung Gwak, Kevin Chen, and Silvio Savarese.
\newblock 3d-r2n2: A unified approach for single and multi-view 3d object
  reconstruction.
\newblock In {\em European conference on computer vision}, pages 628--644.
  Springer, 2016.

\bibitem{danvevcek2022emoca}
Radek Dan{\v{e}}{\v{c}}ek, Michael Black, and Timo Bolkart.
\newblock Emoca: Emotion driven monocular face capture and animation.
\newblock In {\em 2022 IEEE/CVF Conference on Computer Vision and Pattern
  Recognition (CVPR)}, pages 20279--20290. IEEE, 2022.

\bibitem{deng2022depth}
Kangle Deng, Andrew Liu, Jun-Yan Zhu, and Deva Ramanan.
\newblock Depth-supervised nerf: Fewer views and faster training for free.
\newblock In {\em Proceedings of the IEEE/CVF Conference on Computer Vision and
  Pattern Recognition}, pages 12882--12891, 2022.

\bibitem{feng2021learning}
Yao Feng, Haiwen Feng, Michael~J Black, and Timo Bolkart.
\newblock Learning an animatable detailed 3d face model from in-the-wild
  images.
\newblock {\em ACM Transactions on Graphics (ToG)}, 40(4):1--13, 2021.

\bibitem{fridovich2022plenoxels}
Sara Fridovich-Keil, Alex Yu, Matthew Tancik, Qinhong Chen, Benjamin Recht, and
  Angjoo Kanazawa.
\newblock Plenoxels: Radiance fields without neural networks.
\newblock In {\em Proceedings of the IEEE/CVF Conference on Computer Vision and
  Pattern Recognition}, pages 5501--5510, 2022.

\bibitem{gafni2021dynamic}
Guy Gafni, Justus Thies, Michael Zollhofer, and Matthias Nie{\ss}ner.
\newblock Dynamic neural radiance fields for monocular 4d facial avatar
  reconstruction.
\newblock In {\em Proceedings of the IEEE/CVF Conference on Computer Vision and
  Pattern Recognition}, pages 8649--8658, 2021.

\bibitem{gao2022reconstructing}
Xuan Gao, Chenglai Zhong, Jun Xiang, Yang Hong, Yudong Guo, and Juyong Zhang.
\newblock Reconstructing personalized semantic facial nerf models from
  monocular video.
\newblock {\em ACM Transactions on Graphics (TOG)}, 41(6):1--12, 2022.

\bibitem{girdhar2016learning}
Rohit Girdhar, David~F Fouhey, Mikel Rodriguez, and Abhinav Gupta.
\newblock Learning a predictable and generative vector representation for
  objects.
\newblock In {\em European Conference on Computer Vision}, pages 484--499.
  Springer, 2016.

\bibitem{hane2017hierarchical}
Christian H{\"a}ne, Shubham Tulsiani, and Jitendra Malik.
\newblock Hierarchical surface prediction for 3d object reconstruction.
\newblock In {\em 2017 International Conference on 3D Vision (3DV)}, pages
  412--420. IEEE, 2017.

\bibitem{kajiya1984ray}
James~T Kajiya and Brian~P Von~Herzen.
\newblock Ray tracing volume densities.
\newblock {\em ACM SIGGRAPH computer graphics}, 18(3):165--174, 1984.

\bibitem{kania2023hypernerfgan}
Adam Kania, Artur Kasymov, Maciej Zieba, and Przemyslaw Spurek.
\newblock Hypernerfgan: Hypernetwork approach to 3d nerf gan.
\newblock {\em arXiv preprint arXiv:2301.11631}, 2023.

\bibitem{kania2022conerf}
Kacper Kania, Kwang~Moo Yi, Marek Kowalski, Tomasz Trzci{\'n}iski, and Andrea
  Tagliasacchi.
\newblock Conerf: Controllable neural radiance fields.
\newblock In {\em 2022 IEEE/CVF Conference on Computer Vision and Pattern
  Recognition (CVPR)}, pages 18602--18611. IEEE, 2022.

\bibitem{khakhulin2022realistic}
Taras Khakhulin, Vanessa Sklyarova, Victor Lempitsky, and Egor Zakharov.
\newblock Realistic one-shot mesh-based head avatars.
\newblock In {\em Computer Vision--ECCV 2022: 17th European Conference, Tel
  Aviv, Israel, October 23--27, 2022, Proceedings, Part II}, pages 345--362.
  Springer, 2022.

\bibitem{li2017grass}
Jun Li, Kai Xu, Siddhartha Chaudhuri, Ersin Yumer, Hao Zhang, and Leonidas
  Guibas.
\newblock Grass: Generative recursive autoencoders for shape structures.
\newblock {\em ACM Transactions on Graphics (TOG)}, 36(4):1--14, 2017.

\bibitem{li2017learning}
Tianye Li, Timo Bolkart, Michael~J Black, Hao Li, and Javier Romero.
\newblock Learning a model of facial shape and expression from 4d scans.
\newblock {\em ACM Trans. Graph.}, 36(6):194--1, 2017.

\bibitem{liu2020neural}
Lingjie Liu, Jiatao Gu, Kyaw Zaw~Lin, Tat-Seng Chua, and Christian Theobalt.
\newblock Neural sparse voxel fields.
\newblock {\em Advances in Neural Information Processing Systems},
  33:15651--15663, 2020.

\bibitem{LIU2022108774}
Zehua Liu, Yuhe Zhang, Jian Gao, and Shurui Wang.
\newblock Vfmvac: View-filtering-based multi-view aggregating convolution for
  3d shape recognition and retrieval.
\newblock {\em Pattern Recognition}, 129:108774, 2022.

\bibitem{max1995optical}
Nelson Max.
\newblock Optical models for direct volume rendering.
\newblock {\em IEEE Transactions on Visualization and Computer Graphics},
  1(2):99--108, 1995.

\bibitem{mildenhall2020nerf}
Ben Mildenhall, Pratul~P. Srinivasan, Matthew Tancik, Jonathan~T. Barron, Ravi
  Ramamoorthi, and Ren Ng.
\newblock Nerf: Representing scenes as neural radiance fields for view
  synthesis.
\newblock In {\em ECCV}, 2020.

\bibitem{muller2022instant}
Thomas M{\"u}ller, Alex Evans, Christoph Schied, and Alexander Keller.
\newblock Instant neural graphics primitives with a multiresolution hash
  encoding.
\newblock {\em ACM Transactions on Graphics (ToG)}, 41(4):1--15, 2022.

\bibitem{niemeyer2022regnerf}
Michael Niemeyer, Jonathan~T Barron, Ben Mildenhall, Mehdi~SM Sajjadi, Andreas
  Geiger, and Noha Radwan.
\newblock Regnerf: Regularizing neural radiance fields for view synthesis from
  sparse inputs.
\newblock In {\em Proceedings of the IEEE/CVF Conference on Computer Vision and
  Pattern Recognition}, pages 5480--5490, 2022.

\bibitem{roessle2022dense}
Barbara Roessle, Jonathan~T Barron, Ben Mildenhall, Pratul~P Srinivasan, and
  Matthias Nie{\ss}ner.
\newblock Dense depth priors for neural radiance fields from sparse input
  views.
\newblock In {\em Proceedings of the IEEE/CVF Conference on Computer Vision and
  Pattern Recognition}, pages 12892--12901, 2022.

\bibitem{schmid2012tridimensional}
Kendra~K Schmid, David~B Marx, and Ashok Samal.
\newblock Tridimensional regression for comparing and mapping 3d anatomical
  structures.
\newblock {\em Anatomy Research International}, 2012, 2012.

\bibitem{shu2022wasserstein}
Dong~Wook Shu, Sung~Woo Park, and Junseok Kwon.
\newblock Wasserstein distributional harvesting for highly dense 3d point
  clouds.
\newblock {\em Pattern Recognition}, 132:108978, 2022.

\bibitem{sinha2016deep}
Ayan Sinha, Jing Bai, and Karthik Ramani.
\newblock Deep learning 3d shape surfaces using geometry images.
\newblock In {\em European Conference on Computer Vision}, pages 223--240.
  Springer, 2016.

\bibitem{tancik2022block}
Matthew Tancik, Vincent Casser, Xinchen Yan, Sabeek Pradhan, Ben Mildenhall,
  Pratul~P Srinivasan, Jonathan~T Barron, and Henrik Kretzschmar.
\newblock Block-nerf: Scalable large scene neural view synthesis.
\newblock In {\em Proceedings of the IEEE/CVF Conference on Computer Vision and
  Pattern Recognition}, pages 8248--8258, 2022.

\bibitem{thies2016face2face}
Justus Thies, Michael Zollhofer, Marc Stamminger, Christian Theobalt, and
  Matthias Niessner.
\newblock Face2face: Real-time face capture and reenactment of rgb videos.
\newblock In {\em 2016 IEEE Conference on Computer Vision and Pattern
  Recognition (CVPR)}, pages 2387--2395. IEEE Computer Society, 2016.

\bibitem{zimny2022points2nerf}
T Trzci{\'n}ski.
\newblock Points2nerf: Generating neural radiance fields from 3d point cloud.
\newblock {\em arXiv preprint arXiv:2206.01290}, 2022.

\bibitem{verbin2022ref}
Dor Verbin, Peter Hedman, Ben Mildenhall, Todd Zickler, Jonathan~T Barron, and
  Pratul~P Srinivasan.
\newblock Ref-nerf: Structured view-dependent appearance for neural radiance
  fields.
\newblock In {\em 2022 IEEE/CVF Conference on Computer Vision and Pattern
  Recognition (CVPR)}, pages 5481--5490. IEEE, 2022.

\bibitem{wei2021nerfingmvs}
Yi Wei, Shaohui Liu, Yongming Rao, Wang Zhao, Jiwen Lu, and Jie Zhou.
\newblock Nerfingmvs: Guided optimization of neural radiance fields for indoor
  multi-view stereo.
\newblock In {\em Proceedings of the IEEE/CVF International Conference on
  Computer Vision}, pages 5610--5619, 2021.

\bibitem{yang2022continuous}
Fei Yang, Franck Davoine, Huan Wang, and Zhong Jin.
\newblock Continuous conditional random field convolution for point cloud
  segmentation.
\newblock {\em Pattern Recognition}, 122:108357, 2022.

\bibitem{zheng2022avatar}
Yufeng Zheng, Victoria Fern{\'a}ndez~Abrevaya, Marcel B{\"u}hler, Xu Chen,
  Michael~J Black, and Otmar Hilliges.
\newblock Im avatar: Implicit morphable head avatars from videos.
\newblock In {\em 2022 IEEE/CVF Conference on Computer Vision and Pattern
  Recognition (CVPR)}, pages 13535--13545. IEEE, 2022.

\bibitem{zielonka2022instant}
Wojciech Zielonka, Timo Bolkart, and Justus Thies.
\newblock Instant volumetric head avatars.
\newblock {\em arXiv preprint arXiv:2211.12499}, 2022.

\bibitem{zielonka2022towards}
Wojciech Zielonka, Timo Bolkart, and Justus Thies.
\newblock Towards metrical reconstruction of human faces.
\newblock In {\em Computer Vision--ECCV 2022: 17th European Conference, Tel
  Aviv, Israel, October 23--27, 2022, Proceedings, Part XIII}, pages 250--269.
  Springer, 2022.

\end{thebibliography}

}

\end{document}